\documentclass[10pt,twocolumn,letterpaper]{article}

\usepackage[pagenumbers]{iccv} 

\newcommand{\modify}[1]{{\color{black}#1}}

\newcommand{\ourdataset}{CHOrD dataset}
\newcommand{\ourmethod}{CHOrD}

\def\eqref#1{equation~\ref{#1}}
\def\Eqref#1{Equation~\ref{#1}}
\def\Tabref#1{Table~\ref{#1}}

\def\Figref#1{Figure~\ref{#1}}

\definecolor{iccvblue}{rgb}{0.21,0.49,0.74}
\usepackage[pagebackref,breaklinks,colorlinks,allcolors=iccvblue]{hyperref}

\usepackage{CJKutf8}
\usepackage{url}
\usepackage{graphicx}
\usepackage{subcaption}
\usepackage{listings}
\usepackage{color}
\usepackage{booktabs}
\usepackage{pgfplots}
\usepackage{pgfplotstable}
\pgfplotsset{compat=1.17}
\usepackage{colortbl}
\usepackage{soul}
\usepackage{bm}
\usepackage{float}

\def\paperID{9005} 
\def\confName{ICCV}
\def\confYear{2025}

\title{\textcolor{blue}{CHOrD}: Generation of \textcolor{blue}{C}ollision-Free, \textcolor{blue}{H}ouse-Scale, and \textcolor{blue}{Or}ganized \textcolor{blue}{D}igital Twins for 3D Indoor Scenes with Controllable Floor Plans and Optimal Layouts}

\author{Chong Su$^{\dagger, 1}$ \quad Yingbin Fu$^{\dagger, 1}$ \quad Zheyuan Hu$^2$ \quad Jing Yang$^2$ \quad Param Hanji$^2$ \\  Shaojun Wang$^1$ \quad Xuan Zhao$^1$ \quad Cengiz Öztireli$^2$ \quad Fangcheng Zhong$^2$ \\
$^1$KE Holdings Inc. \quad
$^2$Department of Computer Science and Technology, University of Cambridge\\
{\small $^{\dagger}$ indicates equal contribution.}
}

\begin{document}

\twocolumn[{%
\renewcommand\twocolumn[1][]{#1}%
\vspace{-14mm}
\maketitle
\vspace{-10mm}
\begin{center}
    \captionsetup{type=figure}
    \includegraphics[width=\linewidth]{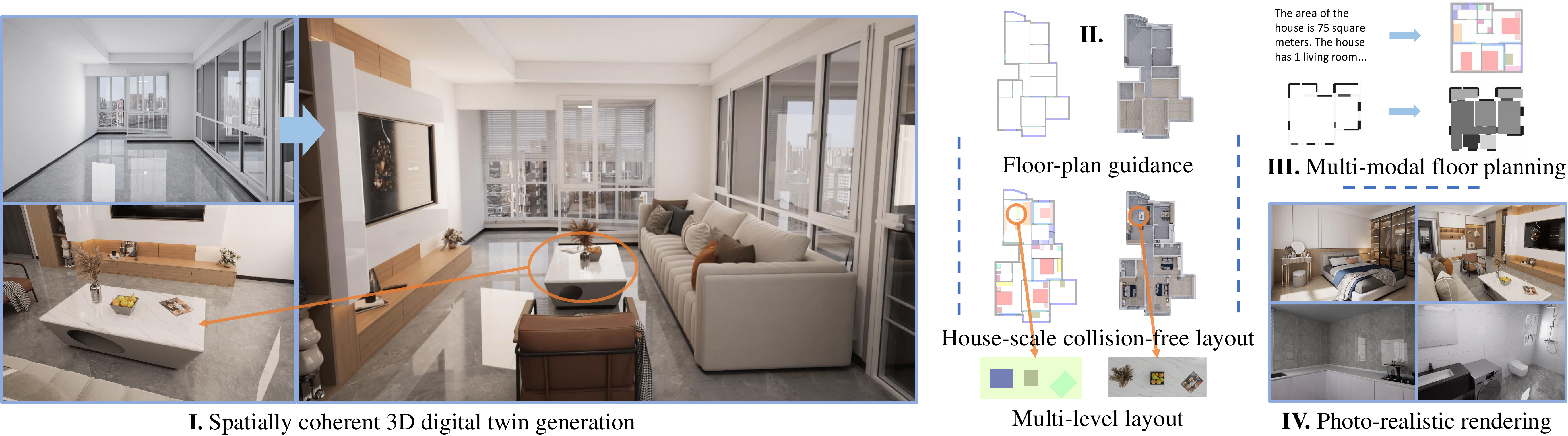}
    \vspace{-4mm}
    \captionof{figure}{ \textit{I)} \ourmethod{} synthesizes realistic and well-structured digital twins for 3D indoor scenes. \textit{II)} \ourmethod{} can be conditioned on complex floor plan structures to generate realistic \textbf{house-wide} layouts while ensuring \textbf{physically plausible}, \textbf{spatially coherent}, and \textbf{collision-free} arrangements. It further introduces a \textbf{hierarchical} data structure that organizes objects not only at the room level but also at finer scales, such as desks and coffee tables. \textit{III, IV)} \ourmethod{} supports controllable floor plans via multimodal inputs, as well as photorealistic, 3D-consistent rendering. These capabilities equip \ourmethod{} with considerable versatility, enabling a broad range of downstream applications.
    }\label{fig:teaser}
\end{center}%
}]

\begin{abstract}
\vspace{-7mm}

%
We introduce \ourmethod{}, a novel framework for scalable synthesis of 3D indoor scenes, designed to create \textbf{house-scale}, \textbf{collision-free}, and \textbf{hierarchically structured} indoor digital twins. In contrast to existing methods that directly synthesize the scene layout as a scene graph or object list, \ourmethod{} incorporates a 2D image-based intermediate layout representation, enabling effective \textbf{prevention of collision artifacts} by successfully capturing them as out-of-distribution (OOD) scenarios during generation.
Furthermore, unlike existing methods, \ourmethod{} is capable of generating scene layouts that \textbf{adhere to complex floor plans} with multi-modal controls, enabling the creation of coherent, house-wide layouts robust to both geometric and semantic variations in room structures. Additionally, we propose a \textbf{novel dataset} with expanded coverage of household items and room configurations, as well as significantly improved data quality. \ourmethod{} demonstrates \textbf{state-of-the-art} performance on both the 3D-FRONT and our proposed datasets, delivering photorealistic, spatially coherent indoor scene synthesis adaptable to arbitrary floor plan variations.
\vspace{-4mm}
\end{abstract}

\section{Introduction}


Generative 3D indoor scene synthesis and digital twin creation~\citep{2002Constraint, 2011Interactive, 2011Make, 2012Example, 2018Human, 2018Deep, 2018GRAINS, 2019Fast, 2019PlanIT, yao2024conditional, 2024FuncScene, 2017Attention, 2020Graph2Plan, 2020SceneFormer, 2021ATISS, 2022LayoutEnhancer, tang2024diffuscene, lin2024instructscene} play increasingly vital roles not only in \emph{creative and technical workflows} such as interior design, architectural planning, and virtual or augmented reality, but also in \emph{advancing embodied AI} by providing scalable simulated environments for training and testing. This approach facilitates rapid prototyping, reduces manual labor, lowers deployment costs, and accelerates iteration. Despite recent advances in neural volumetric representations~\citep{mildenhall2020nerf, kerbl20233d}, classic mesh-based assets remain the predominant 3D digital twin representation in these domains due to superior rendering quality, direct interactivity, and explicit spatial structures. Consequently, existing pipelines~\citep{2018Deep, 2018GRAINS, 2019Fast, 2019PlanIT, 2021ATISS, tang2024diffuscene, lin2024instructscene} primarily follow a procedural generation paradigm, constructing a \emph{scene graph} or \emph{object list} for the scene layout, with each node containing detailed specifications for individual objects. These objects can then be retrieved from a CAD asset dataset and interacted with or rendered by various graphics and physics engines.
Therefore, synthesizing diverse and logical scene layouts has been a core aspect of creating high-quality indoor digital twins.


However, a fundamental limitation of existing methods that construct scene graphs or object lists—either directly by a tabular generative model~\citep{2018Deep, 2018GRAINS, 2019Fast, 2019PlanIT, 2021ATISS, tang2024diffuscene, lin2024instructscene} or by an LLM writing configuration files~\citep{wang2023robogen, yang2023holodeck}—is their \emph{inability} to prevent physically implausible collisions or overlaps between objects, such as a bed intersecting with cabinets, during the generation process. While collision detection can be performed as a post-processing step, it is computationally expensive and lacks scalability. 
Moreover, effectively resolving detected collisions during post-processing remains nontrivial.
Prior work has also attempted to prevent collisions using manually defined rules~\citep{ProcTHOR, yang2023holodeck}, but this approach lacks generalizability to arbitrary scenes and cannot be used to learn desirable layout distributions from data.
Another critical yet frequently overlooked limitation of existing methods is their \emph{restriction} to single-room layout generation~\citep{2018Deep, 2018GRAINS, 2019Fast, 2019PlanIT, 2021ATISS, tang2024diffuscene, lin2024instructscene}, which fails to account for the overall floor plan structure of a house. Since room shapes, sizes, and the arrangement of various room types collectively influence the logical organization of the household layout, existing approaches neglect key spatial relationships that are essential for coherent multi-room designs.
The occurrence of these limitations is not coincidental — neither tabular generative models nor LLMs have the granular spatial understanding needed to distinguish adjacent objects from intersecting ones or to properly incorporate the floor plan geometries into the generation process.

In this paper, we propose \ourmethod{}, a framework designed to comprehensively enhance the spatial coherence of digital twin generation for 3D indoor scenes, as highlighted in Figure~\ref{fig:teaser}. In particular, \ourmethod{} \textit{i)} \emph{prohibits collision artifacts} during the generation process without relying on post-processing, collision detection, or pre-defined rules; \textit{ii)} enables \emph{house-scale layout generation} that adapts to complex geometric and semantic floor plan structures, which are controllable via multi-modal input; and \textit{iii)} supports a \emph{hierarchically structured} scene graph representation that seamlessly integrates into existing pipelines.
Central to our approach is the synthesis of a \emph{2D image-based} layout representation as an intermediate step in the procedural workflow, which can be subsequently converted into a \emph{hierarchical scene graph}, rather than constructing the graph in a single step. Our key insight is that, compared to the graph representation, which is inherently \emph{tabular}, introducing an intermediate 2D layout representation greatly strengthens the spatial perception and reasoning of the generative model. For example, humans can readily spot collisions by examining the top-down view of a layout, whereas simply reviewing a table of bounding boxes does not enable such direct assessment. 
Designers routinely rely on top-down floor plan views to create spatially orderly layouts. In a similar vein, \ourmethod{} successfully captures collision artifacts as \emph{out-of-distribution} (OOD) scenarios, facilitating the empirical elimination of physically implausible collisions during the generation process.
The 2D layout of \ourmethod{} also enables the adaptation of complex floor plan structures via various modalities, a feature rarely available in existing approaches.
We advocate that all graph-based methods for digital twin generation incorporate an intermediate 2D layout for collision avoidance and house-scale organization.

We additionally introduce a novel dataset, referred to as the \ourdataset{}, comprising 9,706 scenes with floor plans and scene layouts, approximately 1.4 times larger than 3D-FRONT~\citep{fu20213d}. Compared to 3D-FRONT, the \ourdataset{} expands household item coverage to 26 super-categories, including items from kitchens, bathrooms, and balconies, addressing gaps in 3D-FRONT, which lacks furnishings in these areas and occasionally leaves living rooms or bedrooms unfurnished. It also resolves common issues found in 3D-FRONT, such as misclassified objects, unrealistic placements, and collisions, providing clean layouts without requiring extensive data cleaning. 

Our contributions are summarized as:
\textit{i)} \textbf{Framework}\footnote{Our codebase and dataset are available in the Supplementary Materials, and will be publicly released upon acceptance.} - 
A novel framework for \emph{house-scale}, \emph{collision-free}, and \emph{hierarchically structured} indoor digital twin creation with multi-modal controllable floor plans;
\textit{ii)} \textbf{Dataset} - A novel dataset of 9,706 scenes with floor plans and scene layouts, 1.4 times larger than 3D-FRONT, with improved item coverage and data quality;
and \textit{iii)} \textbf{Performance} - \ourmethod{}'s state-of-the-art performance on both the 3D-FRONT and proposed datasets, evaluated both qualitatively and quantitatively, particularly in the near-elimination of unreasonable object collisions, a prevalent issue in existing methods.

\begin{figure*}[!ht]
    \centering
    \includegraphics[width=0.98\textwidth]{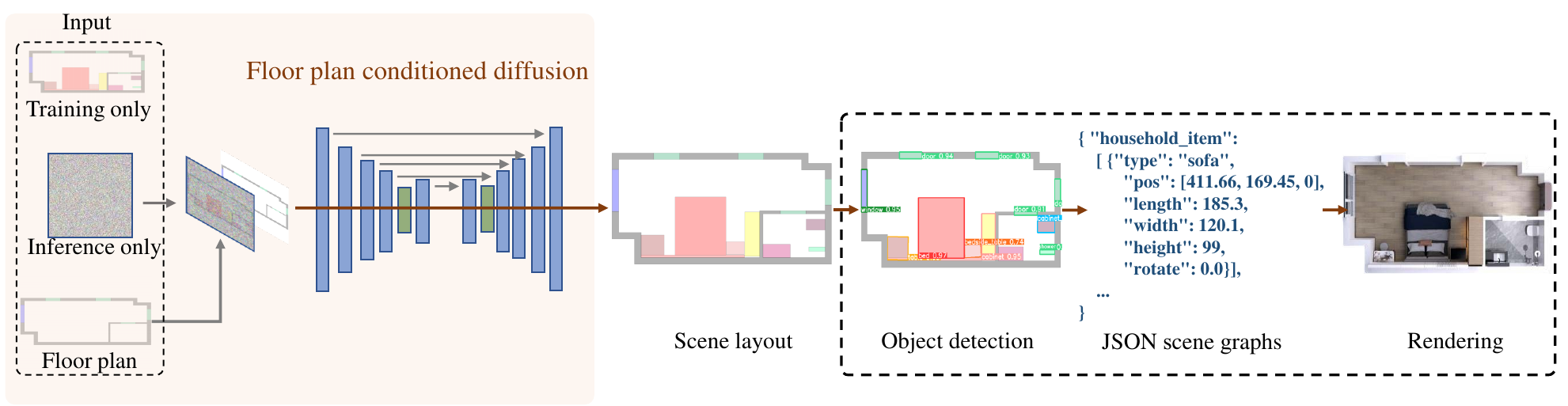}
    \caption{Overview of \ourmethod{}. First, we generate the scene layout using a conditional diffusion model, conditioned on a floor plan image. Next, we apply object detection to identify individual household items and use a structured scene graph to hierarchically organize the spatial relationships between rooms and objects, along with their attributes. Finally, the scene is rendered into photorealistic images.
    }
    \vspace{-4mm}
    \label{fig:overview}
\end{figure*}

\section{Related Work}
Early work employed a rule-based constraint satisfaction formulation to generate 3D room layouts for pre-specified sets of objects \citep{2002Constraint, ProcTHOR}. While allowing for moderate diversity, rule-based methods cannot learn desirable layout distributions from data.
Other approaches optimized cost functions based on interior design principles \citep{2011Interactive} and object-object statistical relationships \citep{2011Make}. The earliest data-driven approach modeled object co-occurrences using a Bayesian network and Gaussian mixtures to capture pairwise spatial relations extracted from 3D scenes \citep{2012Example}.
With the availability of large datasets of 3D environments, such as SUNCG \citep{2017Semantic}, 3D-FRONT \citep{fu20213d}, SUN3D \citep{xiao2013sun3d}, Matterport3D \citep{chang2017matterport3d}, InteriorNet \citep{li2018interiornet}, Structured3D \citep{zheng2020structured3d}, and 3D-FURNITURE \citep{fur20213d}, learning-based approaches have gained popularity. Various methods for indoor scene synthesis have been proposed, including: human-centric probabilistic grammars \citep{2018Human}, Generative Adversarial Networks (GANs) trained on matrix representations of scene objects \citep{2018Deep}, recursive neural networks for sampling 3D scene hierarchies \citep{2018GRAINS}, convolutional neural networks (CNNs) trained on top-down room images \citep{2018Deep, 2019Fast}, spatial prior graph neural networks trained on labeled 3D spatial relationships \citep{2019PlanIT, yao2024conditional}, and Variational Autoencoder (VAE) models trained on top-down functional furniture group images \citep{2024FuncScene}. Additionally, floor plan synthesis approaches have been proposed using graph neural networks \citep{2020Graph2Plan}.
With the development of transformer models \citep{2017Attention}, transformer-based approaches have become increasingly popular. These include floor plan-conditioned furniture synthesis and text-conditioned furniture synthesis using transformer autoregressive models \citep{2020SceneFormer, 2021ATISS}, as well as methods integrating expert knowledge in the form of differentiable scalar functions to guide the generation of more ergonomic layouts \citep{2022LayoutEnhancer}.
Recently, diffusion models \citep{2020Denoising} have demonstrated impressive visual quality in generative tasks, including indoor furniture synthesis \citep{tang2024diffuscene, lin2024instructscene}.
However, these methods primarily synthesize layouts for individual rooms rather than house-scale scenes that consider the overall floor plan structure.
Additionally, unlike recent works \cite{tang2024diffuscene, lin2024instructscene, yang2024physcene}, which directly synthesize the scene graph using a 1D-Unet, our approach employs a 2D-Unet. This enables a better understanding of the spatial relationships between doors, windows, and furniture, thereby more effectively preventing collisions, doorway blockages, and other artifacts. Such artifacts are also present in widely used indoor scene datasets, such as 3D-FRONT~\citep{fu20213d, fur20213d}, requiring substantial effort for data cleaning.

\section{\ourmethod{} Pipeline}

We propose a novel pipeline for 3D-aware indoor scene synthesis and digital twin creation, as depicted in Figure~\ref{fig:overview}. The pipeline starts with a floor plan description—provided as an RGB image—and use a conditional diffusion model to generate a corresponding 2D scene layout. The use of this 2D representation enables us to leverage efficient image encoders for layout generation while effectively distinguishing \emph{natural} and \emph{implausible} object overlaps. Next, we employ automatic object detectors and segmentation maps to identify individual household items and extract a structured scene graph that hierarchically organizes \emph{multi-level} spatial relationships and object attributes. Finally, the 3D scene objects are retrieved accordingly and rendered to produce photorealistic 3D-consistent images, which can also be deployed in a physics engine for simulation.

\subsection{Diffusion-based scene layout generation}
\label{sec:diffusion}
We leverage the recent success of image-based diffusion models~\citep{saharia2022palette,rombach2022high,amit2021segdiff} and frame the problem of generating diverse, realistic indoor scene layouts as a conditional image-to-image translation task, as illustrated in Figure~\ref{fig:overview} (left). Unlike complex scene graphs or tabular formats, natural RGB images serve as a convenient intermediate representation for the layout, easily processed by existing vision tools. Crucially, since RGB images are easy-to-interpret by an appropriate encoder, we can construct a highly effective conditional generative model that accurately captures the data distribution. In 2D images, implausible object collisions are instantly visible and flagged as OOD samples, enabling the model to generate coherent, realistic layouts.

Specifically, given an image of an empty floor plan $\bm{y}$, we train a diffusion model $\epsilon_\theta(\bm{x};\,\bm{y},t)$ to model the conditional distribution of the corresponding layouts $p(\bm{x} \mid \bm{y})$,
where $\epsilon_\theta$ is structured as a 2D U-Net, following \citep{2020Denoising}, with 3 input channels (random noise) and 3 output channels (the predicted layout image). To incorporate floor plan image conditioning, we expand the U-Net input from 3 to 6 channels.
During training, a predetermined noise schedule realizes a Markov chain, yielding the diffused sample $\bm{x_t}(\bm{x},\bm{y},t,\epsilon)$, where $\epsilon \sim \mathcal{N}(\bm{0},\bm{I})$ and $t \sim \mathcal{U}(0,1)$. The loss function is given by the denoising score matching objective~\citep{2020Denoising}:

\begin{equation} \label{eq:floor2layout2}
\mathbb{E}_{(\bm{x}, \bm{y}) \sim p_\textrm{data},\epsilon \sim \mathcal{N}(\bm{0}, \bm{I}),t \sim \mathcal{U}(0,1)} \left[ \left\| \epsilon_\theta(\bm{x};\,\bm{y},t) - \epsilon \right\|^2 \right].
\end{equation}

\subsection{Hierarchical scene graph extraction and object retrieval}
\label{sec:scene-graph}

\begin{figure}
    \centering
    \includegraphics[width=0.98\linewidth]{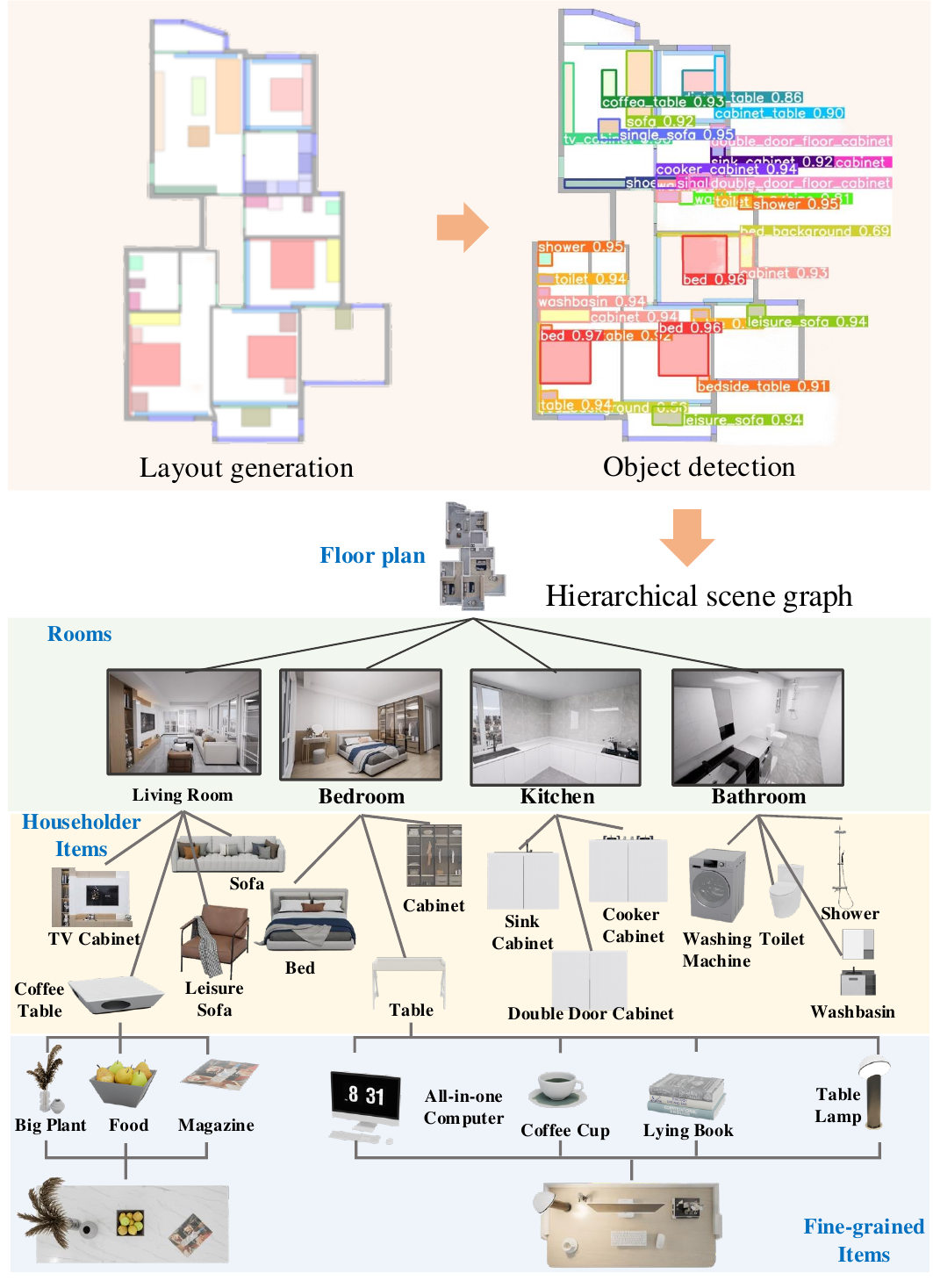}
    \caption{Scene graph extraction and object retrieval.
    }
    \vspace{-4mm}
    \label{fig:scene_graph}
\end{figure}

To generate a scene graph from the candidate layout $\bm{x} \sim p(\bm{x} \mid \bm{y})$, we follow the automatic framework proposed by~\citep{lv2021residential}. As depicted in Figure~\ref{fig:scene_graph}, we start by fine-tuning YOLOv8~\citep{Jocher_Ultralytics_YOLO_2023} to detect the locations and attributes of all objects present in $\bm{x}$. The color of each object uniquely identifies its category from a set of 28 household item categories and 3 floor plan item categories. 
Detailed color schemes are listed in Supplementary \Tabref{tab:furniture}. 
The other relevant attributes are then populated to produce an object list $\mathcal{O} = (\bm{o_1}, \bm{o_2},\dots,\bm{o_n})$, with each node containing object properties such as category, position, orientation, and size.
We employ YOLOv8 to simultaneously obtain the segmentation maps for each room type, including living rooms, bedrooms, kitchens, bathrooms, and balconies.

Given the dimensions and category of each object, we deterministically retrieve an example from a category-specific textured mesh database $\mathcal{D}$\footnote{Note that the selection of this database and its retrieval rules can be flexibly user-specified, enabling custom functionality by incorporating advanced features into the graph nodes, as explored in many prior works~\cite{yang2024physcene, zhai2023commonscenes}. \ourmethod{} can be seamlessly integrated into these pipelines.} such that it has the smallest size difference:
\begin{equation}
    \bm{e_i} = \arg\min_{\bm{e} \in \mathcal{D}}(\left\| o_i^x - e_i^x \right\|^2+\left\| o_i^y- e^y \right\|^2) :o_i^c=e^c, \,\forall \bm{o_i} \in \mathcal{O}.
\end{equation}

The set of retrieved examples $\{\bm{e_1}, \bm{e_2},\dots,\bm{e_n}\}$ constitutes the leaf nodes of the scene graph, as shown in \Figref{fig:scene_graph}. To position the objects in each room and construct the hierarchal scene graph, we utilize the semantic detection and segmentation outputs of household items and rooms from YOLOv8.
We straighten the edges of the room polygons, similar to \citep{lv2021residential}, to reduce uneven lines, and attach doors and windows to these edges, ensuring corrected wall positions that enclose the room. 

Note that this approach enables \ourmethod{} to generate granular, hierarchical spatial layouts in a multi-level \emph{autoregressive} manner. Specifically, we can iteratively apply the conditional diffusion model to generate fine-grained layouts, such as placing objects on a coffee table, as illustrated in Figure~\ref{fig:scene_graph} (bottom). When generating fine-grained layouts, the conditional input for the diffusion model becomes the top-down views of the upper level (\textit{e.g.}, the boundaries of the table) instead of floor plan images. 

The advantages of a hierarchical layout data structure are threefold. First, this structure allows \ourmethod{} to be seamlessly integrated into widely adopted graph-based pipelines to enhance spatial coherence, which we strongly advocate. Second, it facilitates a wide range of downstream tasks, such as intricate robotic spatial understanding and navigation~\citep{Werby-RSS-24}. Finally, this multi-level layout enables \ourmethod{} to also accommodate natural vertical object overlaps, such as placing objects on a desk or coffee table.




\subsection{Multi-modal floor planning}
\label{sec:multimodal}

Apart from the main pipeline, the 2D layout of \ourmethod{} enables additional multimodal controls for the floor plan.
Specifically, we provide two types of controls:

\vspace{-4mm}
\paragraph{Text-conditioned floor planning}
An alternative and convenient way to specify the floor plan is through natural language, especially when floor plan images are not accessible or incompatible with the format accepted by our model. Given the success of text-to-image diffusion models~\citep{ramesh2022hierarchical,saharia2022imagen}, text descriptions provide a viable alternative for floor plan specification, as shown in Figure~\ref{fig:mmfp} (left). 

\vspace{-4mm}
\paragraph{Open-plan-conditioned floor planning}
\ourmethod{} also supports synthesizing floor plans conditioned on an open-plan layout, as shown in Figure~\ref{fig:mmfp} (right). Specifically, given a 2D image of an open-plan layout without room arrangements, \ourmethod{} generates complete floor plans with optimal room separations. This is particularly useful for users looking to modify floor plan structures or synthesize digital twin environments with greater variety.

These controls are considered extended features of \ourmethod{}—the main pipeline functions perfectly without them—but they are made possible largely due to our adoption of a multi-level 2D layout. We anticipate various new features enabled by this approach. Further technical details are provided in Supplementary Section~\ref{sec:supp-implementation-details}.

\subsection{Rendering}
Finally, we convert the structured scene graph into a 3D mesh. The wall and floor materials for each $\bm{e_i}$ are procedurally sampled, while being aware of the rooms to which they belong. To maintain uniform lighting and shadow consistency across the scene, an appropriately sized area light is placed at the center of each room. 
The UE engine \citep{unrealengine} is subsequently utilized to generate photorealistic renderings.

The key advantage of the multi-stage pipeline of \ourmethod{}—which first generates a 2D layout rather than directly synthesizing an object scene graph using a tabular generative model~\citep{tang2024diffuscene, lin2024instructscene} or LLMs~\citep{wang2023robogen, yang2023holodeck}—lies in its enhanced granular spatial understanding, adapting to various floor plan structures and object placements. By leveraging an intermediate 2D layout, \ourmethod{} ensures that the hierarchical spatial relationships between household items are preserved, avoiding common issues such as object overlap, collision, or inconsistencies between object placement and room shapes, as validated in Section~\ref{sec:experiments}.

\section{\ourmethod{} Dataset}
\label{sec:dataset}

We collected a new large-scale dataset, which we refer to as the \ourdataset{}, of indoor scenes with floor plans and scene layouts, comprising a total of 9,706 design schemes, approximately 1.4 times larger than the 3D-FRONT dataset \citep{fu20213d}. This dataset was meticulously created by professional interior designers, stored in JSON format with vectorized data, as exampled in Appendix List~\ref{lst:json}, including wall lines, doors, windows, and household items such as furniture, fixtures, and appliances.
The data description is as follows:

\begin{itemize}
\item \textbf{Rooms}: Represented as enclosed loops of interior wall lines, defined by 2D coordinates.
\item \textbf{Doors, windows, and household items}: Represented as 2D bounding boxes, defined by category, 3D coordinates, orientation, and dimensions (length, width, height).
\end{itemize}

\begin{figure}
    \centering
        \includegraphics[width=0.98\linewidth]{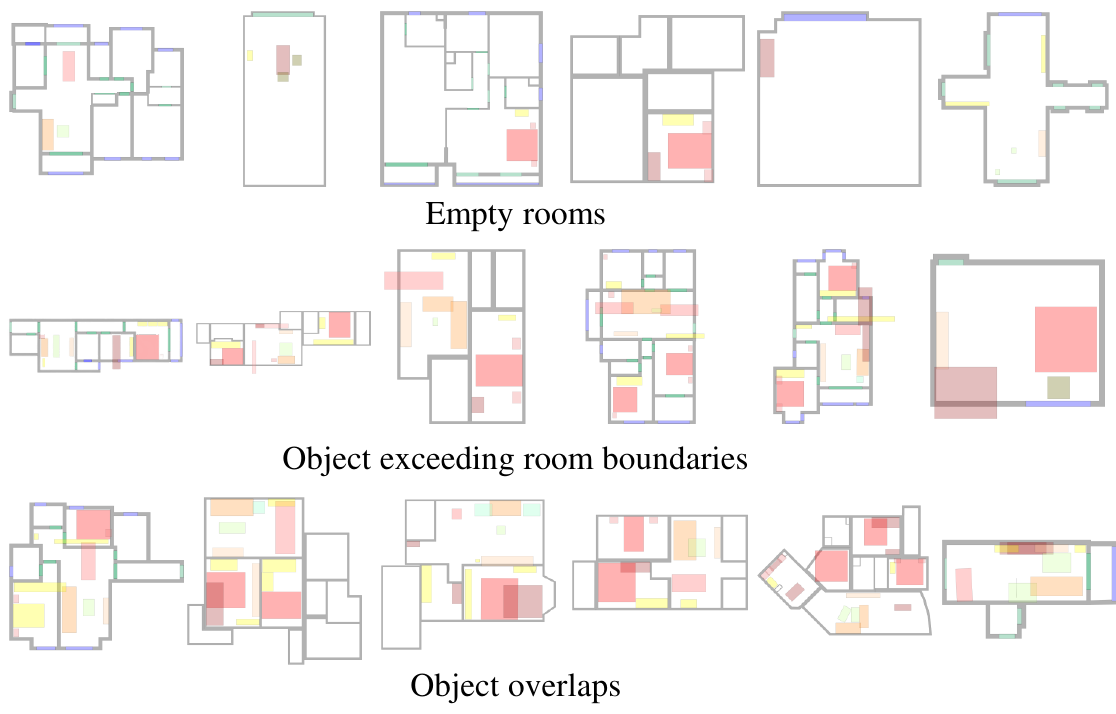}
    \caption{
    Erroneous scenes in 3D-FRONT.
    }
    \vspace{-4mm}
    \label{fig:dataset_compare_visual}
\end{figure}

It is important to note that \ourdataset{} is a 3D \emph{layout} dataset rather than a 3D \emph{asset} dataset. The layout primarily focuses on the geometric characteristics (\textit{e.g.}, bounding boxes) and categorical distinctions among objects.
While \ourdataset{} is currently linked to a small pool of CAD asset models, users are free to retrieve assets from any large public dataset~\citep{3dmodel, fur20213d} to introduce \emph{stylistic variations} of objects if needed. Similarly, 3D-FRONT has been associated with the 3D-FUTURE dataset~\cite{fur20213d} for this purpose.
\ourdataset{} offers several clear advantages over 3D-FRONT:

\vspace{-4mm}
\paragraph{Expanded coverage of household items and room categories} 
While 3D-FRONT provides instance semantic labels for 34 categories and 10 super-categories of household items, its dataset primarily includes objects placed in living rooms, bedrooms, and dining rooms, with no objects for kitchens, bathrooms, or balconies. Consequently, the layouts in 3D-FRONT are consistently devoid of furnishings in these areas, as seen in Figure~\ref{fig:dataset_compare_visual}. 
Our \ourdataset{} fills this gap by offering 26 super-categories of household items, including furniture, fixtures, and appliances, that comprehensively cover living rooms, bedrooms, dining rooms, kitchens, bathrooms, and balconies. 
Our \ourdataset{} not only contains more valid living rooms and bedrooms (each with at least one household item in place), but also includes outfitted kitchens and bathrooms.
A comprehensive statistic of the \ourdataset{} in comparison with 3D-FRONT is detailed in Supplementary Table~\ref{tab:furniture_comparison} and Figure~\ref{fig:statistics}.


\vspace{-4mm}
\paragraph{Improved data quality} 
As frequently reported~\cite{2018Deep,2019Fast,2021ATISS,tang2024diffuscene,lin2024instructscene}, the 3D-FRONT dataset contains erroneous layouts such as empty rooms, unnatural object sizes, misclassified items, and unrealistic object placements (\textit{e.g.}, furniture outside room boundaries, lamps on the floor, blockage of doorways, and overlapping objects), as seen in Figure~\ref{fig:dataset_compare_visual}. Consequently, previous work \citep{2018Deep,2019Fast,2021ATISS,tang2024diffuscene,lin2024instructscene} using 3D-FRONT invested considerable effort in data cleaning, removing numerous layouts with artifacts, which greatly reduced the amount of valid data. 
In contrast, our dataset is ready to use without these artifacts. Supplementary Table~\ref{tab:comparison} and \ref{tab:room_comparison} provide additional details.

\section{Experiments}
\label{sec:experiments}

\begin{figure*}[!ht]
    \centering
    \includegraphics[width=0.98\textwidth]{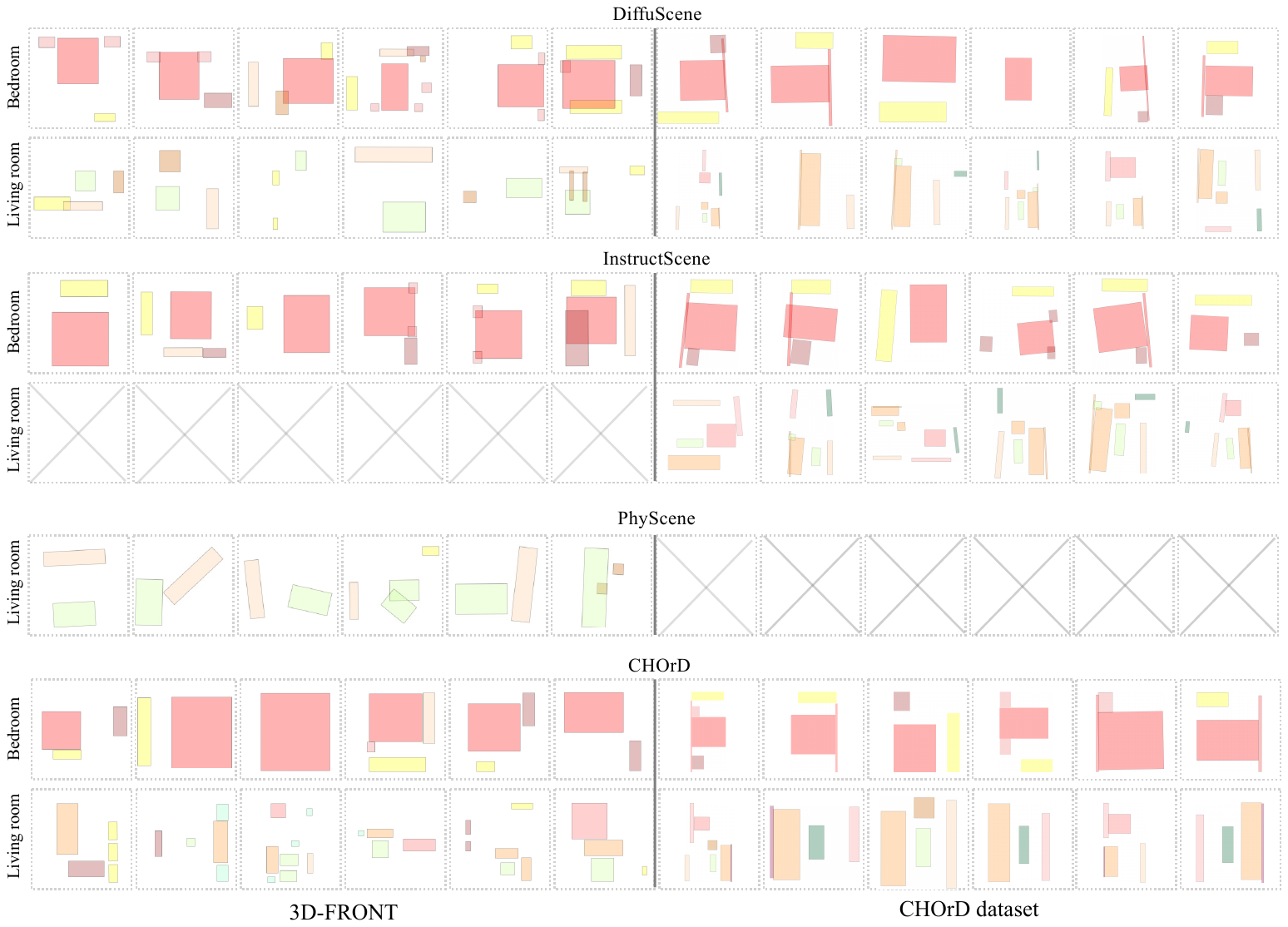}
    \caption{Visualization of synthesized layouts by \ourmethod{}, DiffuScene~\citep{tang2024diffuscene}, InstructScene~\citep{lin2024instructscene}, and PhyScene~\citep{yang2024physcene}.
    All results were randomly selected from an arbitrary batch \emph{without any cherry-picking}. It is evident that only \ourmethod{} produces clean, collision-free layouts, whereas other methods exhibit significant artifacts such as implausible overlapping items, inconsistent orientations, or missing objects.}
    \vspace{-4mm}
    \label{fig:image_grid_}

\end{figure*}

\begin{table*}[h]
    \centering
    \vspace{4mm}
    \resizebox{0.95\linewidth}{!}{
    \begin{tabular}{lc|cccc|cccc|cccc}
        \toprule
        & Dataset & \multicolumn{4}{c}{Bedroom} & \multicolumn{4}{|c}{Living Room} & \multicolumn{4}{|c}{Entire House} \\
        \cline{3-14}
        & & FID$\downarrow$ & KID$\downarrow$ & POR$\downarrow$ & PIoU$\downarrow$ & FID$\downarrow$ & KID$\downarrow$ & POR$\downarrow$ & PIoU$\downarrow$ & FID$\downarrow$ & KID$\downarrow$ & POR$\downarrow$ & PIoU$\downarrow$ \\
        \midrule
        DiffuScene & 3D-FRONT & 15.91 & 0.04 & 0.1632 & 0.0152 & 45.89 & 0.034 & 0.05 & 0.012 & - & - & - & -\\
        InstructScene & 3D-FRONT & 22.35 & 0.02 & 0.2039 & 0.0088 & - & -& - &- & - & -& - & -\\
        \modify{PhyScene} & 3D-FRONT & - & - & - & - & 117.29 & 0.119 & 0.389 &0.0134 & - & -& - & -\\
        \ourmethod{} (ours) & 3D-FRONT & \textbf{14.78} & \textbf{0.008} & \textbf{0.0766} & \textbf{0.0013} & \textbf{24.15} & \textbf{0.018} & \textbf{0.0207} & \textbf{0.0015} & 11.51 & 0.01 & 0.0130 &  0.0005 \\
        \midrule
        DiffuScene & \ourdataset{} & 37.16 & 0.03 & 0.1922 & 0.0038 & 29.97 & 0.02 & 0.0707  & 0.0028 & - & - & - & - \\
        InstructScene & \ourdataset{} & 48.59 & 0.05 & 0.3010 & 0.0092 & 46.05 & 0.04 & 0.0908  & 0.0037 & - & - & - & - \\
        \ourmethod{} (ours) & \ourdataset{} & \textbf{21.86} & \textbf{0.02} & \textbf{0.1049} & \textbf{0.0025} & \textbf{26.69} & \textbf{0.02} & \textbf{0.0485} & \textbf{0.0021} & 29.97 & 0.039 & 0.0125 & 0.0007\\
        \bottomrule
    \end{tabular}}
    \caption{Quantitative evaluation of \ourmethod{} against prior approaches, demonstrating superior performance across all metrics and datasets. 
    }
    \vspace{-4mm}
    \label{tab:layout_comparison}
\end{table*}

We conducted several experiments to assess the performance of \ourmethod{} on structured layout synthesis and compare it with prior work.
We particularly evaluate the effectiveness of \ourmethod{} in eliminating collision artifacts by assessing its ability to capture these scenarios as out-of-distribution samples.
Next, we demonstrate the versatility of \ourmethod{} in several extended tasks, including fine-grained layout synthesis, multi-model floor planning, and photorealistic rendering.

\subsection{Floor plan-conditioned synthesis}
\label{sec:eval}

\paragraph{Implementation}
We trained \ourmethod{} on four RTX 8000 GPUs with a batch size of 4 for 400 epochs. The initial learning rate was set to 1e-4, with a decay factor of 0.1 every 100 epochs. For the diffusion process, we followed the default configuration of DDPM~\citep{2020Denoising}, where noise intensity gradually increases from 0 to 1 over 1000 time steps. For the object detection process, we followed the default configuration of YOLOv8~\cite{Jocher_Ultralytics_YOLO_2023}.
Further implementation details can be found in Supplementary Section~\ref{sec:supp-implementation-details}.

\vspace{-4mm}
\paragraph{Datasets}
We compare \ourmethod{} with baseline methods on both the 3D-FRONT dataset~\cite{fu20213d} and the proposed \ourdataset{}. 
The 3D-FRONT dataset consists of 6,813 scenes, of which 4,847 were retained after a cleaning process that excluded layouts lacking furniture, containing objects extending beyond room boundaries, or exhibiting collisions. Prior works~\citep{2018Deep,2019Fast,2021ATISS,tang2024diffuscene,lin2024instructscene} have applied similar data filtering to remove erroneous scenes from 3D-FRONT due to various artifacts, as discussed in Section~\ref{sec:dataset}.
The \ourdataset{} comprises 9,706 scenes and is ready for use without the need for data cleaning or preprocessing. We use 80\% of the dataset for training and 20\% for testing.

\vspace{-4mm}
\paragraph{Baselines}
We compare \ourmethod{} with the latest work DiffuScene~\citep{tang2024diffuscene}, InstructScene~\citep{lin2024instructscene}, and PhyScene~\citep{yang2024physcene}, all aiming to synthesize 3D indoor scenes with optimized layouts.
Note that DiffuScene, InstructScene, and PhyScene are all \emph{unable} to synthesize house-scale layouts but individual categories of rooms.

For evaluation on the 3D-FRONT dataset, we used the official pre-trained checkpoints of these methods to ensure their optimal performance. Specifically, we used the checkpoint from the DiffuScene unconditional model to generate top-down views of object arrangements in bedrooms and living rooms at a resolution of $256\times 256$, matching the image size generated by our diffusion model. For InstructScene, we similarly used the checkpoint from the unconditional model to generate bedroom views at the same resolution. InstructScene did not release unconditional model checkpoints for living rooms. For PhyScene, we used their checkpoint from the floorplan-conditioned model to generate living room layouts. PhyScene did not release model checkpoints for bedrooms. To ensure fairness in the comparison, the object categories generated by DiffuScene, InstructScene, and PhyScene were remapped to our categorization, as detailed in Supplementary Table~\ref{tab:furniture_map}. 
For evaluation on the \ourdataset{}, we re-trained the unconditional models of DiffuScene and InstructScene on living rooms and bedrooms using their default training configurations. PhyScene did not release its training code.

\begin{figure}
    \centering
    \vspace{-2.5mm}
        \includegraphics[width=1\linewidth]{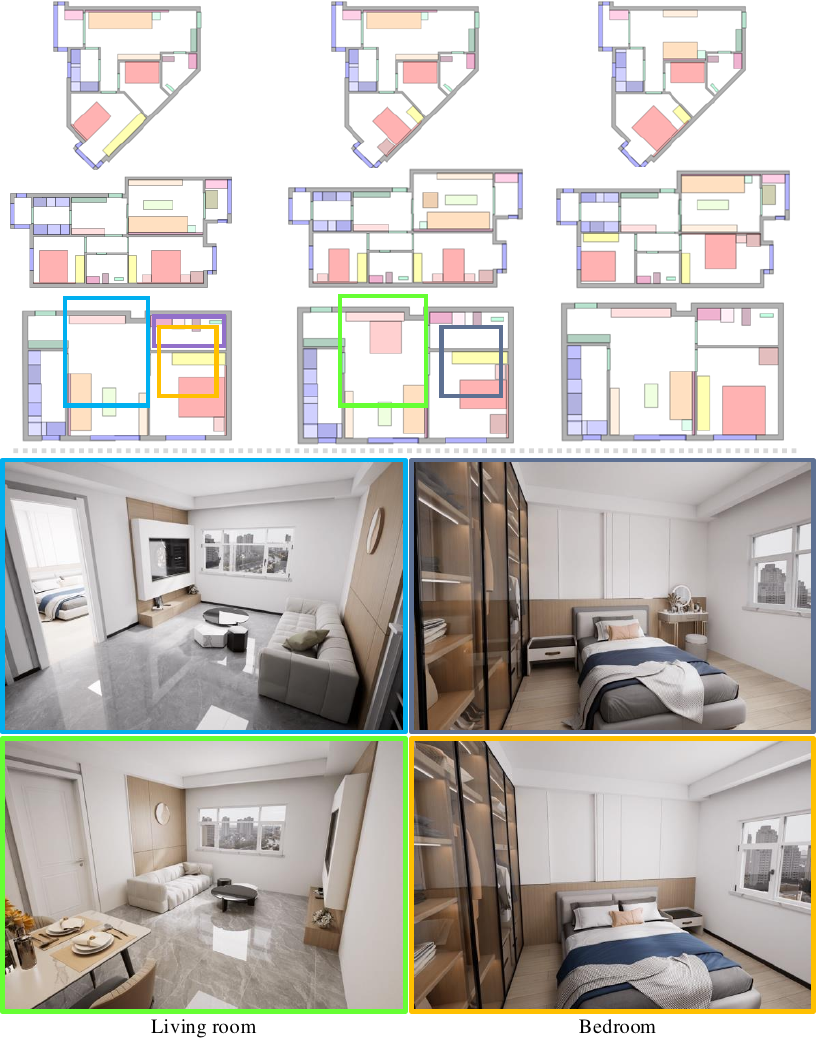}
    \caption{ \textbf{Top} - Visualization of three diverse layouts (columns) synthesized by \ourmethod{} for each of the three floor plans (rows). \ourmethod{} is robust to irregular and slanted room shapes.
    \textbf{Bottom} - Photorealistic rendering of living rooms and bedrooms with identical camera positions and floor plans, highlighting layout diversity.
    The correspondence between the layout on the top and the rendering on the bottom is indicated by matching colored frames.
    \vspace{-4.5mm}
    }
    \label{fig:floor_cond_render}
\end{figure}

\vspace{-4mm}
\paragraph{Results}
We present the qualitative evaluation of all methods in Figure~\ref{fig:image_grid_}, with all results randomly selected without cherry-picking. \ourmethod{} effectively synthesizes diverse, spatially coherent, and collision-free layouts, while other methods produce significant artifacts, including physically implausible object collisions, inconsistent object orientations, and missing objects, greatly limiting their practical applicability. Moreover, unlike \ourmethod{}, these methods cannot generate house-scale layouts covering all rooms. Figure~\ref{fig:floor_cond_render} illustrates house-scale layouts synthesized by \ourmethod{}, as well as photorealistic renderings. Additional results are available in Supplementary Materials.
Notably, \ourmethod{} can generate diverse 2D layouts from the same floor plan, despite \ourdataset{} containing only one layout per plan.

We present the quantitative evaluation of all methods in Table~\ref{tab:layout_comparison}.
Following previous work~\cite{lin2024instructscene, tang2024diffuscene, yang2024physcene}, we use Frechet Inception Distance (FID) \citep{heusel2017gans} and Kernel Inception Distance (KID) \citep{binkowski2018demystifying} to assess the quality and diversity of synthesized layout images. Additionally, we compute two metrics to evaluate 2D bounding box collisions in synthesized layouts: Pairwise Overlap Ratio (POR), which quantifies the proportion of intersecting object pairs relative to the total number of pairs, and Pairwise Intersection over Union (PIoU), which measures the ratio of the intersecting area between two objects to the combined area of their union. The average values for these metrics are obtained by first computing per-scene values, followed by applying the arithmetic mean.
\ourmethod{} consistently achieves state-of-the-art performance across all metrics and datasets.

\vspace{-4mm}
\paragraph{Collision as OOD samples}
In diffusion models, the training loss is computed as the reconstruction error of the data given the noise, serving as an approximation of the negative log-likelihood (NLL). 
After adequate training, if a sample has a high training loss and, consequently, a high NLL, it is most likely an out-of-distribution (OOD) sample within the learned distribution that is \emph{improbable} to be generated during sampling.
To validate the effectiveness of \ourmethod{} in preventing implausible collision artifacts by recognizing them as OOD samples during inference, we computed the training loss for clean 3D-FRONT layout samples devoid of collisions and for a set of 400 3D-FRONT samples with the largest PIoU values. The loss was calculated by adding noise to the samples at timesteps ranging from 900 to 1000, measuring the mean squared error between the true and predicted noise, and averaging the results over 100 iterations.
The results indicate an average loss of $5.37 \times 10^{-5}$ for the clean samples and $7.10 \times 10^{-5}$ for the samples with collisions,
a significant 32.22\% difference explaining the efficacy of \ourmethod{} in identifying and prohibiting unnatural object collisions as OODs.

\begin{figure}
    \centering
    \includegraphics[width=0.98\linewidth]{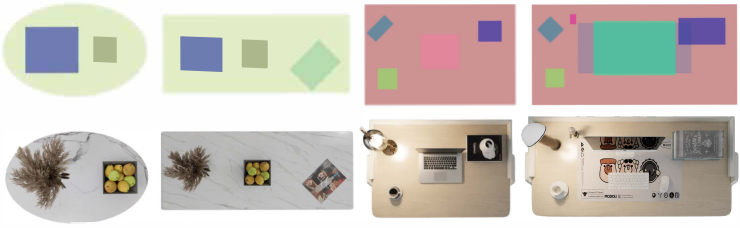}
    \caption{
    Fine-grained coffee table and desk layouts that accommodate natural vertical object overlaps. 
    The computer setup in the right column, consisting of a monitor, keyboard, and mouse, was modeled as a single object placed on the mat. 
    }
    \vspace{-4mm}
    \label{fig:table_decorations}
\end{figure}

\subsection{Fine-grained layout synthesis}
\label{sec:fine-layout}

As discussed in Section~\ref{sec:scene-graph}, the multi-level graph structure enables \ourmethod{} to synthesize fine-grained layouts such as placing objects on a coffee table. This can be achieved by iteratively applying the conditional diffusion model, except that the floor plan image conditioning is replaced by an image indicating the boundaries of upper-level items.

\vspace{-4mm}
\paragraph{Dataset and implementation}
Since neither the 3D-FRONT nor \ourmethod{} datasets contain fine-grained layouts for this task, we additionally collected a small dataset of object placements on common household items such as dining tables, coffee tables, and desks. We recorded the object categories, positions, orientations, and sizes, as well as their bounding boxes, and generated top-view images of their layouts. Object categorization and their color schemes are detailed in Supplementary Table~\ref{table:table_decorations}. The objects were drawn proportionally to their absolute sizes, with the maximum drawing area fixed at 2-meter squares. We adhered to the same training procedures as detailed in Section~\ref{sec:eval}.

\vspace{-4mm}
\paragraph{Results} We present exemplar results in Figure~\ref{fig:table_decorations}. \ourmethod{} enables two mechanisms that simultaneously prevent implausible object collisions while allowing natural vertical overlaps.
First, as discussed in Section~\ref{sec:scene-graph}, the iterative multi-level layout generation allows fine-grained objects to be placed on upper levels, such as a computer on a desk. Second, some vertical overlaps do not exhibit clear hierarchical relationships, such as an object partially resting on a desk mat. In this scenario, we directly train the diffusion model with RGB images containing vertical overlaps, enabling it to generate plausible layouts with natural vertical overlaps while preventing unreasonable ones. The unique color assigned to each object guides the 2D diffusion model in distinguishing permissible overlaps from invalid ones.

Due to the limited availability of naturally occurring partially overlapped objects, we demonstrate this feature only at the fine-grained level. Figure~\ref{fig:table_decorations} illustrates both scenarios.


\subsection{Multi-modal floor planning}

\paragraph{Text-conditioned floor planning}
For text conditioning, we parse the JSON file of each scene in the \ourdataset{} to extract the total area, room count, and categories to generate the corresponding textual description. 

\vspace{-4mm}
\paragraph{Open-plan-conditioned floor planning}
We use the \ourdataset{} to generate open-plan layouts and floor plans with proper room arrangements as grayscale images, with different colors representing room types. 

Both experiments followed the same training procedures as detailed in Section~\ref{sec:eval}. We present exemplar results in Figure~\ref{fig:mmfp} and more in Supplementary Figure~\ref{fig:text_cond_gen_exp}, \ref{fig:frame_gen}.

\begin{figure}
    \centering
    \vspace{-2.5mm}
    \includegraphics[width=.98\linewidth]{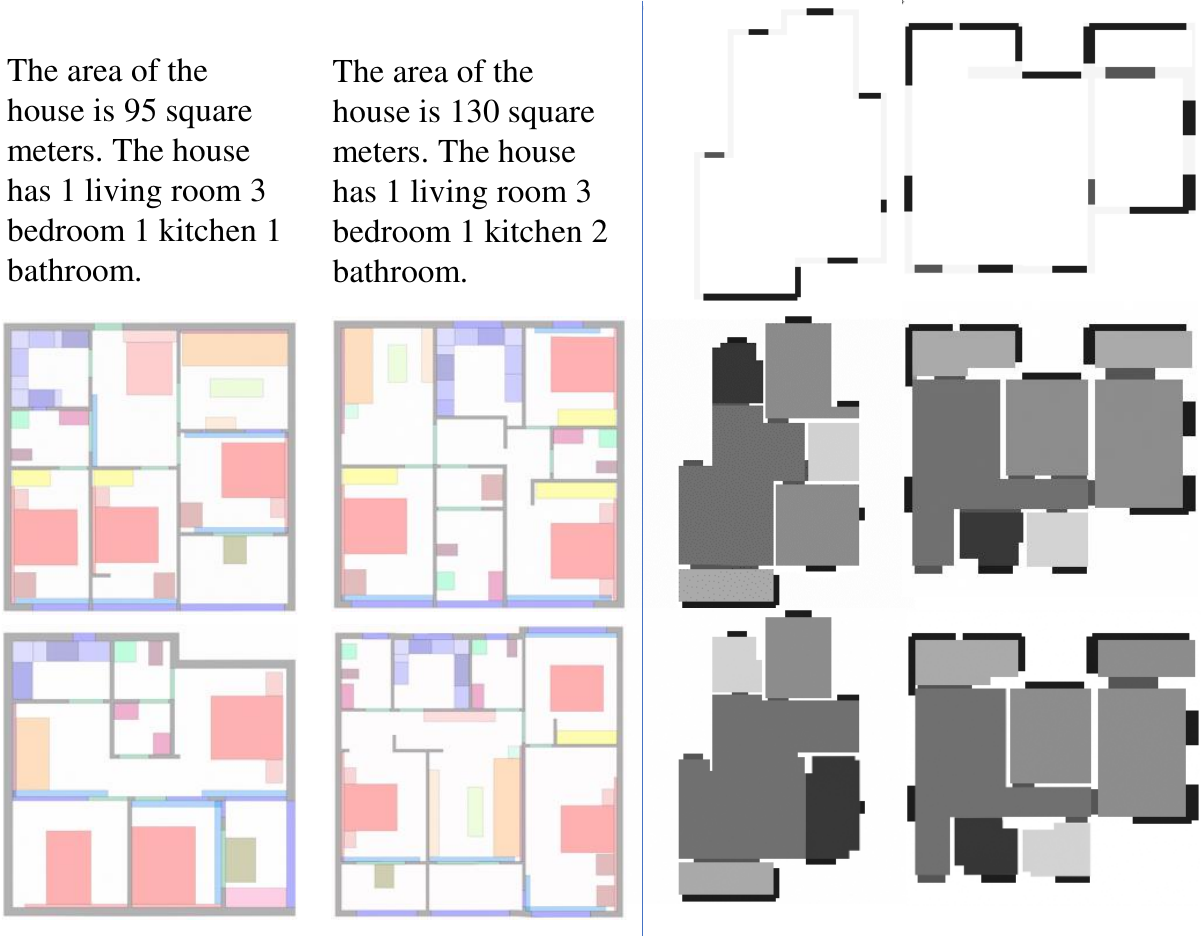}
    \caption{Multi-modal floor planning.}
    \vspace{-4.5mm}
    \label{fig:mmfp}
\end{figure}


\section{Discussions and Summary}
\label{sec:conclusion}
In this paper, we propose a novel framework that employs a 2D image-based intermediate layout representation to ensure house-scale, spatially coherent, collision-free, and hierarchically structured digital twins for indoor 3D scenes. The success of \ourmethod{} hinges on its comprehensively enhanced spatial understanding compared to existing solutions, such as tabular generative models or LLMs, which struggle to meet these objectives.
Notably, \ourmethod{} prevents implausible object collisions while allowing natural vertical overlaps, demonstrating considerable robustness. \ourmethod{} can seamlessly integrate into existing graph-based pipelines for digital twin creation, enabling photorealistic rendering and physics simulation for various downstream tasks.

\vspace{-4mm}
\paragraph{Limitations} \ourmethod{} did not explore stylistic control of individual objects or text-guided object placement, as has been explored by prior works~\cite{lin2024instructscene, tang2024diffuscene}. However, as \ourmethod{} can be integrated into these pipelines, we leave these features for future work. In extremely rare cases, YOLOv8 failed to detect precise bounding boxes, leading to misoriented objects or minor collisions despite the layout images being axis-aligned and collision-free. This can be readily addressed with more training data, thanks to the strong scalability of \ourmethod{}, as evidenced in Supplementary Section~\ref{sec:additional-res}. 
With the same amount of training data, \ourmethod{} outperforms prior work by a significant margin.

{
    \small
    \bibliographystyle{ieeenat_fullname}
    \bibliography{main}
}

\clearpage
\twocolumn[{
\begin{center}
    {\Large \textbf{\textcolor{blue}{CHOrD}: Generation of \textcolor{blue}{C}ollision-Free, \textcolor{blue}{H}ouse-Scale, and \textcolor{blue}{Or}ganized \textcolor{blue}{D}igital Twins for 3D Indoor Scenes with Controllable Floor Plans and Optimal Layouts}} \\[2ex]
    {\Large Supplementary Materials}
\end{center}
}]
\section*{}
\renewcommand{\thesubsection}{\Alph{subsection}}

\subsection{Additional \ourmethod{} Technical Details}
\label{sec:supp-implementation-details}

\subsubsection{\ourmethod{} Inference Efficiency}
On a single RTX 8000 GPU, the diffusion model inference takes approximately 18 seconds (this can be potentially reduced with advanced diffusion solvers); YOLO object detection takes around 40 milliseconds; 3D model matching and scene construction take about 100 milliseconds; and rendering, performed using the UE engine, takes approximately 30 seconds for a 2K image and around 120 seconds for a 4K image. While rendering is the most time-consuming module, it is an independent component that can be flexibly replaced with any real-time rasterization-based renderer when efficiency is a priority. We chose a ray-tracing-based renderer for photorealistic quality. 

\subsubsection{Text-conditioned floor planning}

\begin{figure}
    \centering
    \includegraphics[width=\linewidth]{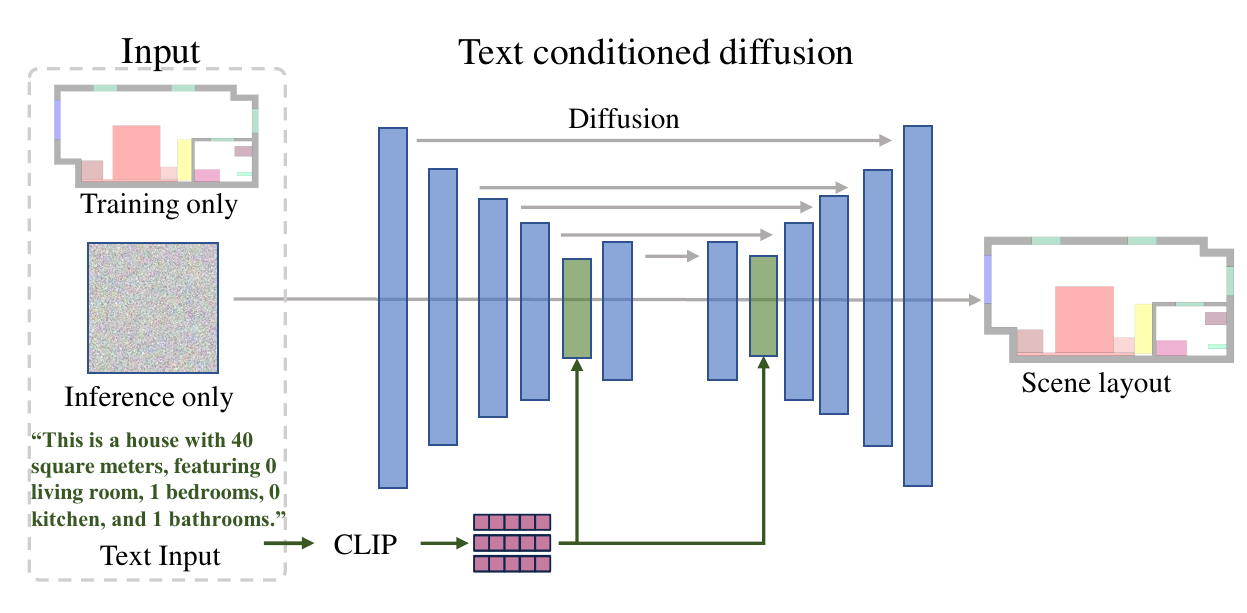}
    \caption{
    Text-conditioned diffusion model.
    }
    \label{fig:text_cond}
\end{figure}

As discussed in Section~\ref{sec:multimodal}, text conditioning serves as a viable alternative for layout generation when floor plans images are not available.
We use a text-conditioned diffusion model for this purpose, as illustrated in Figure~\ref{fig:text_cond}.
We generate a fixed-size conditioning vector $\bm{y_c}$ by passing the text input through a CLIP encoder~\citep{radford2021learning}.
Semi-structured text is particularly effective for this task (\textit{e.g.}, “This is a 40-square-meter flat with 0 living rooms, 1 bedroom, 0 kitchens, and 1 bathroom.”).
The resulting CLIP embedding serves as the conditional variable for the diffusion model, guiding the generation of scene layouts based on high-level semantic information encoded in the text description. This allows for more intuitive control over the layout generation by leveraging natural language as an additional input modality.
The text-based model is trained using the same loss as \Eqref{eq:floor2layout2}, with the conditioning variable being the CLIP embedding $\bm{y_c}$ instead of the floor plan image $\bm{y}$.
Conditioning is introduced through a cross-attention layer~\citep{vaswani2017attention} near the UNet bottleneck. 

\subsubsection{Open-plan-conditioned floor planning}

\begin{figure}
    \centering
    \includegraphics[width=\linewidth]{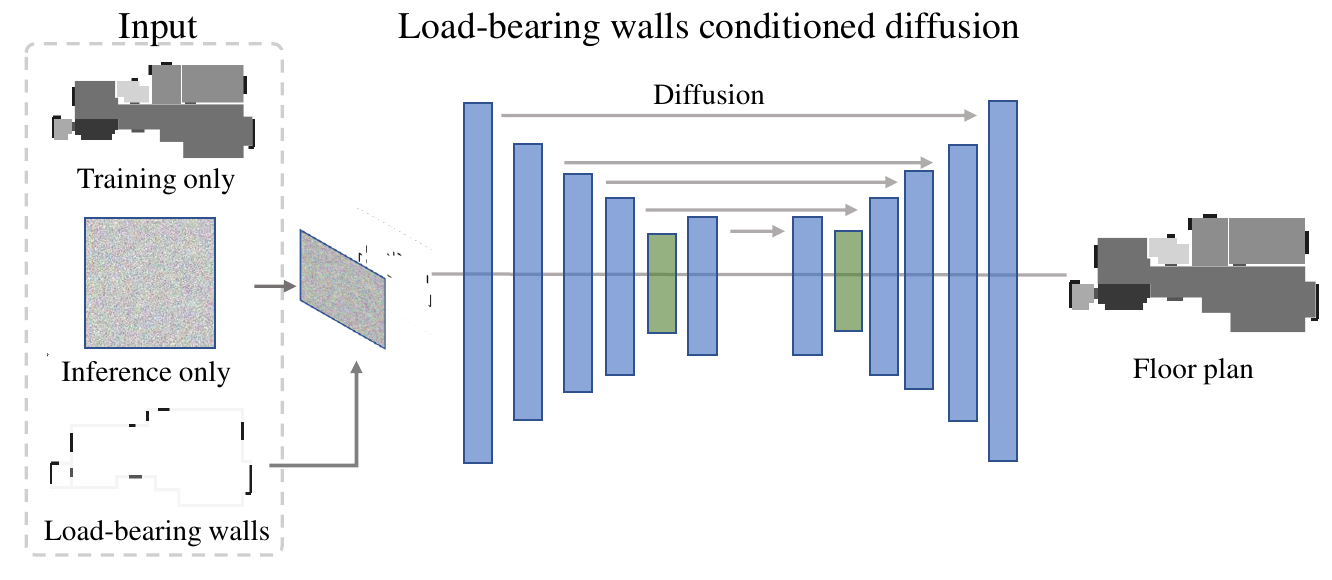}
    \caption{
    Open-plan-conditioned diffusion model.
    }
    \label{fig:wall_cond}
\end{figure}

As illustrated in Figure~\ref{fig:wall_cond}, the open-plan-conditioned diffusion model shares the same architecture as the floor plan-conditioned diffusion model detailed in Section~\ref{sec:diffusion}, except that this model takes an open-plan figure as input and generates a structured floor plan with optimal room arrangements. The generated floor plan can then serve as input to the floor plan-conditioned diffusion model. In other words, open-plan-conditioned floor planning functions as an optional preprocessing step before the \ourmethod{} main pipeline.

\begin{table}[tbp]
    \centering
    \resizebox{\linewidth}{!}{
    \begin{tabular}{|l|l||l|l|}
        \hline
        \textbf{Category} & \textbf{Color} & \textbf{Category} & \textbf{Color} \\
        \hline
        Bed & \textcolor[HTML]{FF0000}{FF0000} & Cabinet & \textcolor[HTML]{FFFF00}{FFFF00} \\
        \hline
        Bed Background & \textcolor[HTML]{FF3333}{FF3333} & Bedside Table & \textcolor[HTML]{F08080}{F08080} \\
        \hline
        Table & \textcolor[HTML]{A52A2A}{A52A2A} & Leisure Sofa & \textcolor[HTML]{666600}{666600} \\
        \hline
        Sofa & \textcolor[HTML]{FF9933}{FF9933} & TV Cabinet & \textcolor[HTML]{FFCC99}{FFCC99} \\
        \hline
        Sofa Background & \textcolor[HTML]{99004C}{99004C} & Coffea Table & \textcolor[HTML]{CCFF99}{CCFF99} \\
        \hline
        Dining Cabinet & \textcolor[HTML]{FF9999}{FF9999} & Shoe Cabinet & \textcolor[HTML]{006633}{006633} \\
        \hline
        Single Sofa & \textcolor[HTML]{CC6600}{CC6600} & Dining Table & \textcolor[HTML]{FF6666}{FF6666} \\
        \hline
        Side Coffea Table & \textcolor[HTML]{99FFCC}{99FFCC} & Single Door Floor Cabinet & \textcolor[HTML]{9999FF}{9999FF} \\
        \hline
        Double Door Floor Cabinet & \textcolor[HTML]{6666FF}{6666FF} & Cooker Cabinet & \textcolor[HTML]{000099}{000099} \\
        \hline
        Sink Cabinet & \textcolor[HTML]{0000CC}{0000CC} & Electrical FLoor Cabinet & \textcolor[HTML]{3333FF}{3333FF} \\
        \hline
        Refrigerator & \textcolor[HTML]{006666}{006666} & Shower & \textcolor[HTML]{33FF99}{33FF99} \\
        \hline
        Toilet & \textcolor[HTML]{660033}{660033} & Washbasin & \textcolor[HTML]{CC0066}{CC0066} \\
        \hline
        Washing Machine & \textcolor[HTML]{FFCCE5}{FFCCE5} & Washing Set & \textcolor[HTML]{FF66B2}{FF66B2} \\
        \hline
        Wall & \textcolor[HTML]{000000}{000000} & Door & \textcolor[HTML]{139C5A}{139C5A} \\
        \hline
        Window & \textcolor[HTML]{0000FF}{0000FF} & & \\
        \hline
    \end{tabular}}
    \caption{Scene layout items and corresponding color schemes, with the opacity level set to 0.3.}
    \label{tab:furniture}
\end{table}

\begin{table}[h]
\centering
\small
\begin{tabular}{|l|l|l|l|}
\hline
\textbf{Category} & \textbf{Color} & \textbf{Category} & \textbf{Color} \\ \hline
Bedside Table & \textcolor[HTML]{F08080}{F08080} & Table & \textcolor[HTML]{A52A2A}{A52A2A} \\ \hline
Coffea Table & \textcolor[HTML]{CCFF99}{CCFF99} & Side Coffea Table & \textcolor[HTML]{99FFCC}{99FFCC} \\ \hline
Dining Table & \textcolor[HTML]{FF6666}{FF6666} & Lying Book & \textcolor[HTML]{0000FF}{0000FF} \\ \hline
Standing Book & \textcolor[HTML]{FFFFAA}{FFFFAA} & Magazine & \textcolor[HTML]{7FFFAA}{7FFFAA} \\ \hline
All-in-one Computer & \textcolor[HTML]{00FFAA}{00FFAA} & Laptop & \textcolor[HTML]{FF7FAA}{FF7FAA} \\ \hline
Big Mouse Pad & \textcolor[HTML]{7F7FAA}{7F7FAA} & Table Lamp & \textcolor[HTML]{007FAA}{007FAA} \\ \hline
Small Ornament & \textcolor[HTML]{FF00AA}{FF00AA} & Pen Holder & \textcolor[HTML]{7F00AA}{7F00AA} \\ \hline
Big Plant & \textcolor[HTML]{0000AA}{0000AA} & Small Plant & \textcolor[HTML]{FFFF55}{FFFF55} \\ \hline
Coffee Cup & \textcolor[HTML]{7FFF55}{7FFF55} & Electronic & \textcolor[HTML]{FF0000}{FF0000} \\ \hline
Photo Frame & \textcolor[HTML]{FF7F55}{FF7F55} & Food & \textcolor[HTML]{7F7F55}{7F7F55} \\ \hline
Dinner Set & \textcolor[HTML]{FFFF00}{FFFF00} & Drinks & \textcolor[HTML]{7F7F00}{7F7F00} \\ \hline
\end{tabular}
\caption{Fine-grained items and corresponding color schemes. 
}
\label{table:table_decorations}
\end{table}

\subsection{Additional \ourmethod{} Dataset Details}

\begin{CJK*}{UTF8}{gbsn}

\begin{figure}[ht]
    \centering
    \begin{subfigure}[b]{0.33\textwidth}
        \begin{tikzpicture}
        \begin{axis}[
            xbar,
            xmin=0,
            xmax=25000,
            width=1.0\textwidth,
            height=8cm,
            enlarge y limits=0.1,
            xmajorgrids=true,
            xminorgrids=true,
            xlabel={Occurence},
            symbolic y coords={Washing Set, Washing Machine, Washbasin, Toilet, Shower, Refrigerator, Electrical FLoor Cabinet, Sink Cabinet, Cooker Cabinet, Double Door Cabinet, Single Door Cabinet, Side Coffee Table, Dining Table, Single Sofa, Shoe Cabinet, Dining Cabinet, Coffee Table, Sofa Background, TV Cabinet, Sofa, Leisure Sofa, Table, Bedside Table, Bed Background, Cabinet, Bed},
            ytick=data,
            nodes near coords,
            bar width=1pt,
            tick label style={font=\tiny},
            legend style={at={(0.95,0.05)}, anchor=south east, font=\tiny},
            label style={font=\tiny},
            point meta=explicit symbolic,
            scaled x ticks=false, 
        ]
        \addplot coordinates {(10620,Bed) (17649,Cabinet) (0,Bed Background) (14333,Bedside Table) (8318,Table) (237,Leisure Sofa) (6564,Sofa) (6821,TV Cabinet) (0,Sofa Background) (6565,Coffee Table) (1169,Dining Cabinet) (0,Shoe Cabinet) (0,Single Sofa) (5822,Dining Table) (6300,Side Coffee Table) (0,Single Door Cabinet) (0,Double Door Cabinet) (0,Cooker Cabinet) (0,Sink Cabinet) (0,Electrical FLoor Cabinet) (0,Refrigerator) (0,Shower) (0,Toilet) (0,Washbasin) (0,Washing Machine) (0,Washing Set)};
        \addplot coordinates {(24354,Bed) (19365,Cabinet) (16619,Bed Background) (10439,Bedside Table) (9359,Table) (8953,Leisure Sofa) (8430,Sofa) (7935,TV Cabinet) (7019,Sofa Background) (7005,Coffee Table) (5368,Dining Cabinet) (4817,Shoe Cabinet) (3939,Single Sofa) (3444,Dining Table) (4195,Side Coffee Table) (15889,Single Door Cabinet) (14156,Double Door Cabinet) (6904,Cooker Cabinet) (6773,Sink Cabinet) (2081,Electrical FLoor Cabinet) (1307,Refrigerator) (15174,Shower) (15026,Toilet) (12517,Washbasin) (970,Washing Machine) (4153,Washing Set)};
        \legend{3D-FRONT, \ourdataset{}}
        \end{axis}
        \end{tikzpicture}
        \caption{Distribution of household item occurrences per super-category.}
        \label{fig:furniture_comparison}
    \end{subfigure}
    \hfill
    \begin{subfigure}[b]{0.46\textwidth}
        \begin{tikzpicture}
        \begin{axis}[
            xbar,
            xmin=0,
            xmax=1200,
            width=1.0\textwidth,
            height=8cm,
            enlarge y limits=0.1,
            xmajorgrids=true,
            xminorgrids=true,
            xlabel={Occurence},
            ylabel={Room Counts per House},
            ylabel style={at={(axis description cs:-0.1,.55)}, anchor=south}, 
            symbolic y coords={1, 2, 3, 4, 5, 6, 7, 8, 9, 10, 11, 12, 13, 14, 15, 16, 17, 18, 19, 20, 21, 22, 23, 24, 28},
            ytick={1, 3, 5, 7, 9, 11, 13, 15, 17, 19, 21, 23, 28}, 
            bar width=1pt,
            tick label style={font=\tiny},
            label style={font=\tiny},
            legend style={at={(0.98,0.05)}, anchor=south east, font=\tiny}
        ]
        \addplot coordinates {(0,1) (226,2) (201,3) (315,4) (542,5) (781,6) (946,7) (1002,8) (753,9) (615,10) (455,11) (325,12) (225,13) (136,14) (81,15) (73,16) (57,17) (22,18) (13,19) (8,20) (12,21) (4,22) (0,23) (5,24) (1,28)};
        \addplot coordinates {(30,1) (127,2) (151,3) (230,4) (397,5) (683,6) (862,7) (992,8) (1094,9) (1168,10) (1032,11) (951,12) (730,13) (495,14) (297,15) (206,16) (105,17) (67,18) (37,19) (23,20) (14,21) (8,22) (3,23) (3,24) (1,28)};
        \legend{3D-FRONT, \ourdataset{}}
        \end{axis}
        \end{tikzpicture}
        \caption{Distribution of room counts per house, with an average of 9.78 and a total of 94,964 counts.}
        \label{fig:subfig2}
    \end{subfigure}
    \caption{Statistics of the \ourdataset{} in comparison with 3D-FRONT. 
    }
    \label{fig:statistics}
\end{figure}
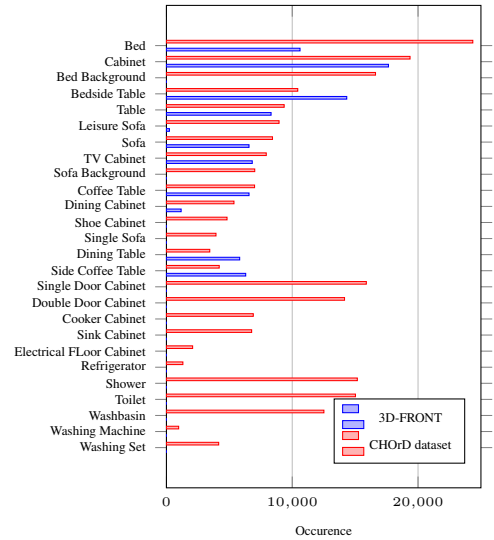
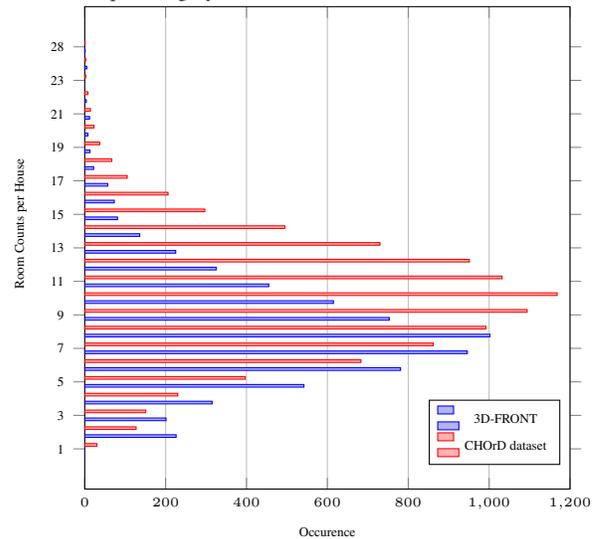

An example \ourmethod{} data stored in JSON format is shown in List~\ref{lst:json}. A comprehensive statistic of the \ourdataset{} in comparison with 3D-FRONT is detailed in Table~\ref{tab:furniture_comparison}, \ref{tab:comparison}, \ref{tab:room_comparison}, and Figure~\ref{fig:statistics}.

\begin{lstlisting}[caption={Example JSON data format}, label={lst:json}]
{
  "rooms": [
    {
      "roomId": "D5F19A0446724E",
      "roomName": "living", # inner room
      "roomType": 1, 
      "wallPoints": [
        [171.65, 241.5], 
        [651.66, 241.5], 
        ...] # 2d coords
    },
    {
      "roomId": "D5F19A044672",
      "roomName": "out_room",
      "roomType": 0,
      "wallPoints": [
        [171.65, 241.5], 
        [651.66, 241.5], 
        ...] # 2d coords
    }],
  "windowsDoors": [
    {
      "type": "door",
      "pos": [717.32, 737.0, 0],
      # box center position x,y;
      # height to floor z
      "length": 95,
      "width": 12,
      "height": 210,
      "rotate": 100 
      # roate angle in degree
    },
    {
      "type": "window",
      "pos": [657.66, 945.12, 90],
      "length": 153.75,
      "width": 12,
      "height": 110,
      "rotate": 90.0
    }
    ],
  "furniture": [
    # 3d bounding box data,
    # same with windows and doors
    {
      "type": "coffee_table",
      "pos": [569.91, 1844.75, 0],
      "length": 76.0,
      "width": 94.0,
      "height": 99,
      "rotate": 180.0
    },
    {
      "type": "sofa",
      "pos": [411.66, 169.45, 0],
      "length": 185.3,
      "width": 120.1,
      "height": 99,
      "rotate": 0.0
    }
    ]
}
\end{lstlisting}

\begin{table}[h]
    \centering
    \small
    \begin{tabular}{|l|l|}
        \hline
        \textbf{Item} & \textbf{Category} \\ \hline
        Nightstand & bedside table \\ \hline
        Wardrobe & cabinet \\ \hline
        Three-Seat / Multi-seat Sofa & sofa \\ \hline
        Dining Table & dining table \\ \hline
        Coffee Table & coffee table \\ \hline
        Loveseat Sofa & sofa \\ \hline
        Children Cabinet & cabinet \\ \hline
        Drawer Chest / Corner cabinet & cabinet \\ \hline
        King-size Bed & bed \\ \hline
        TV Stand & tv cabinet \\ \hline
        Sideboard / Side Cabinet / Console & dining cabinet \\ \hline
        Lazy Sofa & leisure\_sofa \\ \hline
        Dressing Table & table \\ \hline
        Wine Cabinet & dining cabinet \\ \hline
        L-shaped Sofa & sofa \\ \hline
        Corner/Side Table & side coffee table \\ \hline
        Bookcase / jewelry Armoire & cabinet \\ \hline
        Kids Bed & bed \\ \hline
        Sideboard / Side Cabinet / Console Table & table \\ \hline
        Bed Frame & bed \\ \hline
        Shoe Cabinet & shoe cabinet \\ \hline
        Three-Seat / Multi-person sofa & sofa \\ \hline
        Double Bed & bed \\ \hline
        Bunk Bed & bed \\ \hline
        Desk & table \\ \hline
        Two-seat Sofa & sofa \\ \hline
        Tea Table & coffee table \\ \hline
        Couch Bed & bed \\ \hline
        Single bed & bed \\ \hline
        Chaise Longue Sofa & sofa \\ \hline
        U-shaped Sofa & sofa \\ \hline
    \end{tabular}
    \caption{3D-FRONT furniture items and remapped categories.}
    \label{tab:furniture_map}
\end{table}

\begin{table}[h]
    \centering
    \scriptsize
    \begin{tabular}{l l l l}
        \toprule
        Room & Furniture & 3D-FRONT & \ourdataset{} (ours) \\ 
        \midrule
        Bedroom & Bed & 10620 & 24354 \\
        & Cabinet & 17649 & 19365 \\
        & Bed Background & 0 & 16619 \\
        & Bedside Table & 14333 & 10439 \\
        & Table & 8318 & 9359 \\
        \midrule
        Living Room & Leisure Sofa & 237 & 8953 \\
        & Sofa & 6564 & 8430 \\
        & TV Cabinet & 6821 & 7935 \\
        & Sofa Background & 0 & 7019 \\
        & Coffee Table & 6565 & 7005 \\
        & Dining Cabinet & 1169 & 5368 \\
        & Shoe Cabinet & 0 & 4817 \\
        & Single Sofa & 0 & 3939 \\
        & Dining Table & 5822 & 3444 \\
        & Side Coffee Table & 6300 & 4195 \\
        \midrule
        Kitchen & Single Door Cabinet & 0 & 15889 \\
        & Double Door Cabinet & 0 & 14156 \\
        & Cooker Cabinet & 0 & 6904 \\
        & Sink Cabinet & 0 & 6773 \\
        & Electrical Cabinet & 0 & 2081 \\
        & Refrigerator & 0 & 1307 \\
        \midrule
        Bathroom & Shower & 0 & 15174 \\
        & Toilet & 0 & 15026 \\
        & Washbasin & 0 & 12517 \\
        & Washing Machine & 0 & 970 \\
        \midrule
        Balcony & Washing Machine Cabinet & 0 & 4153 \\
        \bottomrule
    \end{tabular}
    \caption{Comparison of furniture occurrences between 3D-FRONT and \ourdataset{}.}
    \label{tab:furniture_comparison}
   \end{table}

    \begin{table}[h]
    \centering
    \begin{tabular}{lccc}
        \toprule
        & Empty Room Rate & POR & PIoU \\
        \midrule
        3D-FRONT & 0.5906 & 0.0361 & 0.2547 \\
        \ourdataset{} (ours) & 0.2902 & 0.0044 & 0.0018 \\
        \bottomrule
    \end{tabular}
    \caption{
    Comparison of data quality statistics between 3D-FRONT and \ourdataset{}.
    }
    \label{tab:comparison}
  \end{table}
  
\begin{table}[h]
\centering
\scriptsize
\begin{tabular}{lccccc}
    \toprule
    & Living & Bedroom & Kitchen & Bathroom & Balcony \\
    \midrule
    3D-FRONT & 1813 & 4041 & 0 & 0 & 0  \\
    \ourdataset{} (ours) & 15115 & 40983 & 8262 & 16351 & 8262 \\
    \bottomrule
\end{tabular}
\caption{Comparison of non-empty room statistics between 3D-FRONT and \ourdataset{}.}
\label{tab:room_comparison}
\end{table}




\subsection{Additional Results}
\begin{figure}
    \centering
    \includegraphics[width=\linewidth]{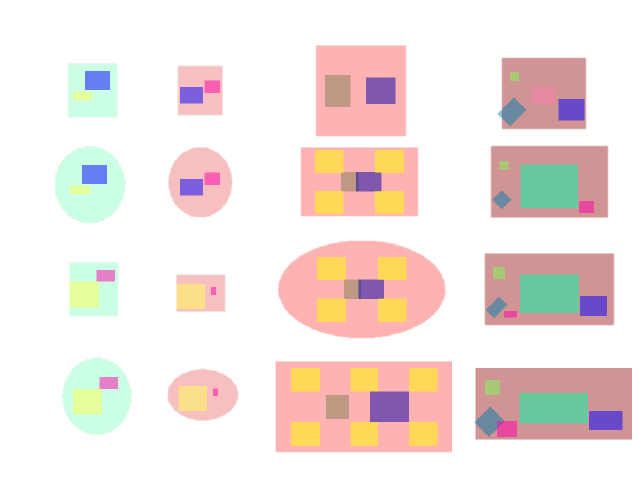}
    \caption{Fine-grained layout synthesis.}
    \vspace{-4mm}
    \label{fig:table_grids}
\end{figure}

\begin{figure}
    \centering
    \includegraphics[width=\linewidth]{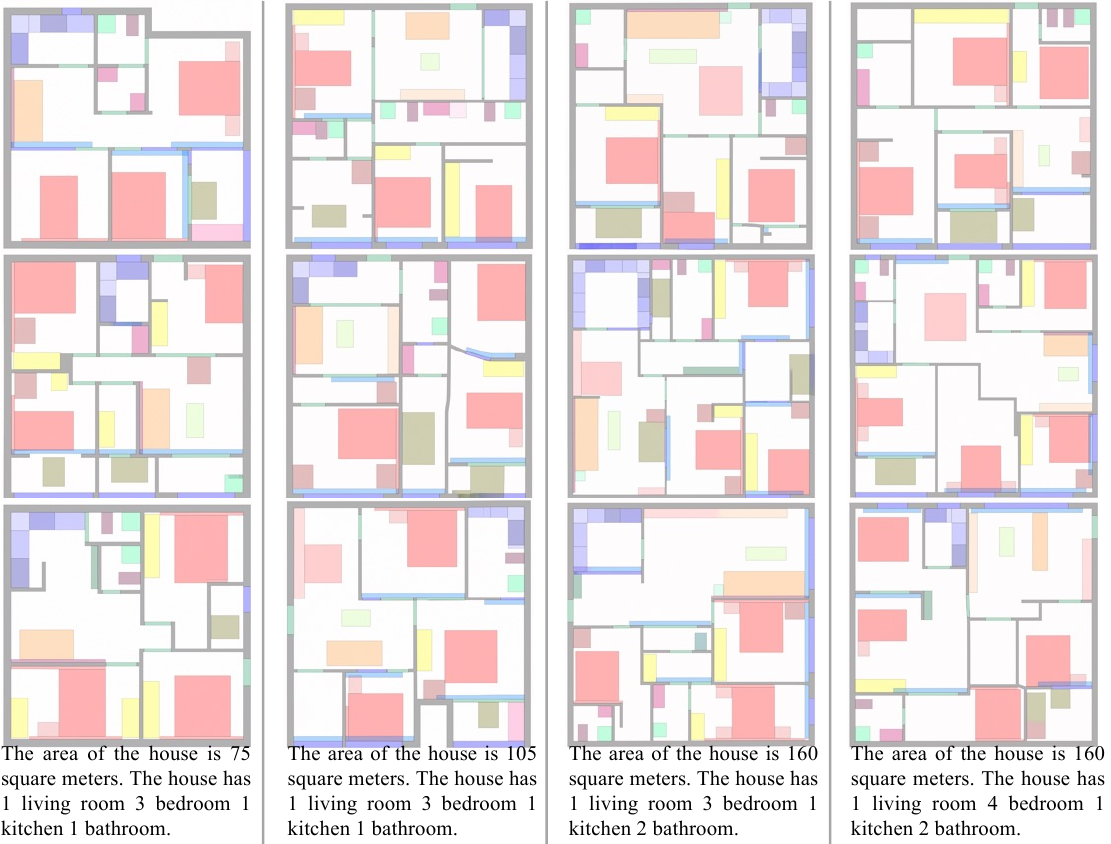}
    \caption{Visualization of text-to-layout generation by \ourmethod{} trained on our \ourdataset. Floor plans of different room sizes all fill the entire canvas, with the wall thickness set to 24 cm for all scenes. Hence, the room size can be inferred from the thickness of the gray walls, which is consistent with the raw training data.}
    \vspace{-4mm}
    \label{fig:text_cond_gen_exp}
\end{figure}

\begin{figure}
    \centering
    \begin{tabular}{c|c|c|c}
        \includegraphics[width=0.2\linewidth]{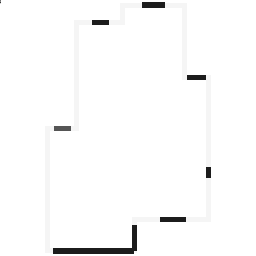} &
        \includegraphics[width=0.2\linewidth]{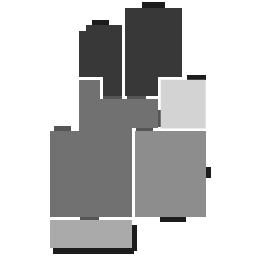} &
        \includegraphics[width=0.2\linewidth]{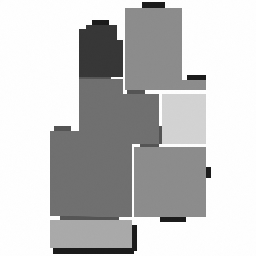}&
        \includegraphics[width=0.2\linewidth]{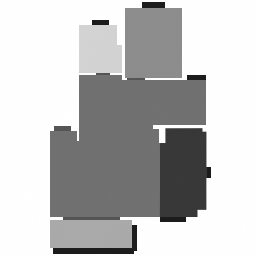} \\
        
        \includegraphics[width=0.2\linewidth]{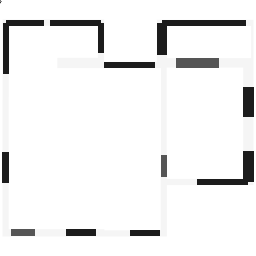} &
        \includegraphics[width=0.2\linewidth]{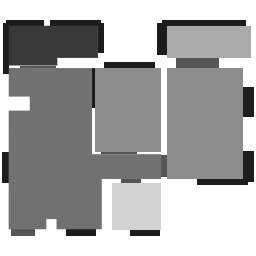} &
        \includegraphics[width=0.2\linewidth]{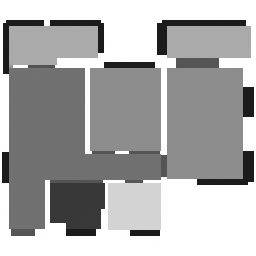}&
        \includegraphics[width=0.2\linewidth]{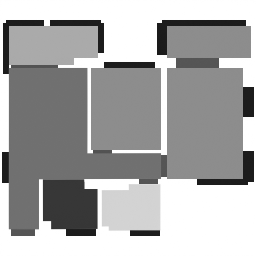} \\
        
        \includegraphics[width=0.2\linewidth]{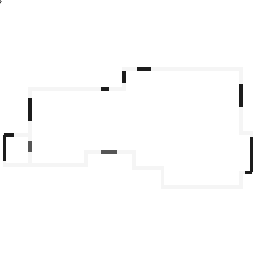} &
        \includegraphics[width=0.2\linewidth]{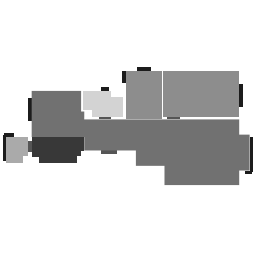} &
        \includegraphics[width=0.2\linewidth]{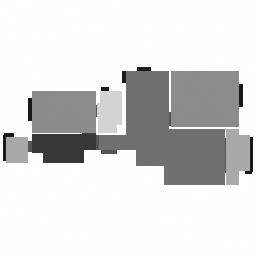}&
        \includegraphics[width=0.2\linewidth]{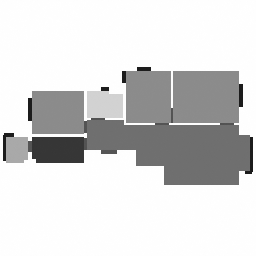} \\
        
        \includegraphics[width=0.2\linewidth]{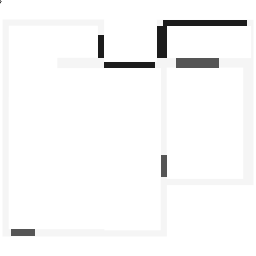} &
        \includegraphics[width=0.2\linewidth]{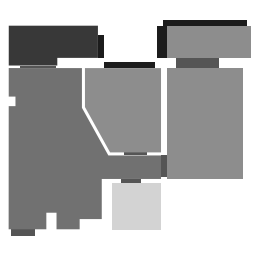} &
        \includegraphics[width=0.2\linewidth]{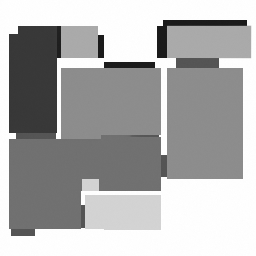}&
        \includegraphics[width=0.2\linewidth]{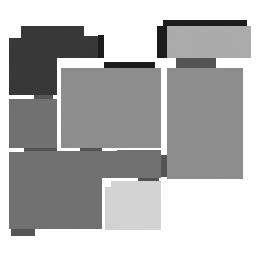} \\
        Condition & Real & Prediction & Prediction \\
    \end{tabular}
    \caption{Open-plan-conditioned floor planning.} 
    \label{fig:frame_gen}
\end{figure}

\begin{figure}[htbp]
    \centering
    \scalebox{0.8}{
        \begin{tabular}{ccc}
            \begin{subfigure}{0.3\textwidth}
                \centering
                \includegraphics[width=\linewidth]{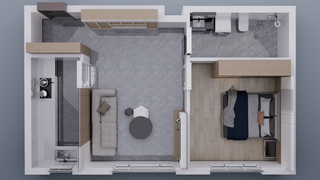}
            \end{subfigure} &
            \begin{subfigure}{0.3\textwidth}
                \centering
                \includegraphics[width=\linewidth]{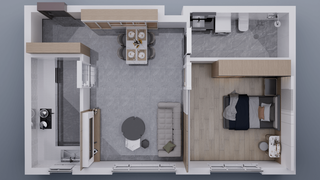}
            \end{subfigure} &
            \begin{subfigure}{0.3\textwidth}
                \centering
                \includegraphics[width=\linewidth]{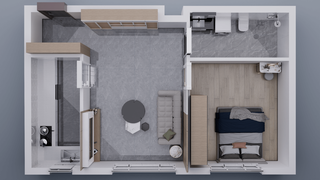}
            \end{subfigure} \\
            
            \begin{subfigure}{0.3\textwidth}
                \centering
                \includegraphics[width=\linewidth]{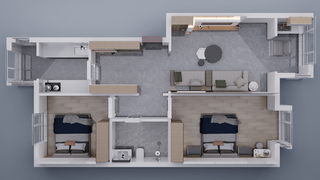}
            \end{subfigure} &
            \begin{subfigure}{0.3\textwidth}
                \centering
                \includegraphics[width=\linewidth]{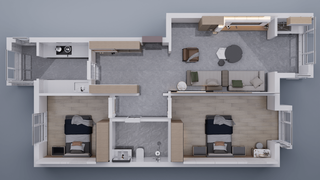}
            \end{subfigure} &
            \begin{subfigure}{0.3\textwidth}
                \centering
                \includegraphics[width=\linewidth]{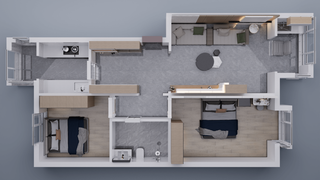}
            \end{subfigure} \\
            
            \begin{subfigure}{0.3\textwidth}
                \centering
                \includegraphics[width=\linewidth]{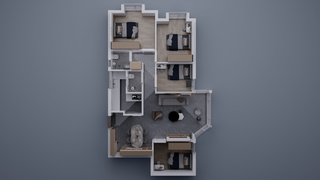}
            \end{subfigure} &
            \begin{subfigure}{0.3\textwidth}
                \centering
                \includegraphics[width=\linewidth]{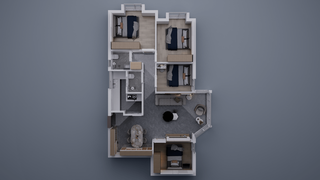}
            \end{subfigure} &
            \begin{subfigure}{0.3\textwidth}
                \centering
                \includegraphics[width=\linewidth]{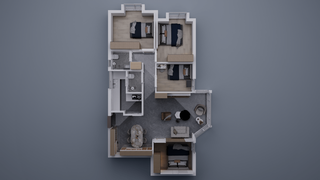}
            \end{subfigure} \\
            \begin{subfigure}{0.3\textwidth}
                \centering
                \includegraphics[width=\linewidth]{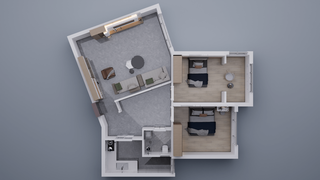}
            \end{subfigure} &
            \begin{subfigure}{0.3\textwidth}
                \centering
                \includegraphics[width=\linewidth]{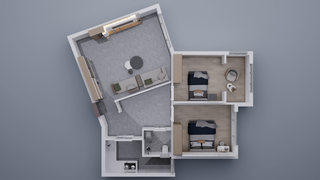}
            \end{subfigure} &
            \begin{subfigure}{0.3\textwidth}
                \centering
                \includegraphics[width=\linewidth]{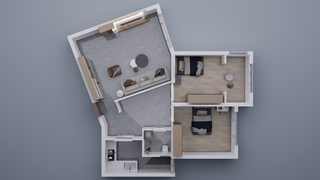}
            \end{subfigure} \\
            \begin{subfigure}{0.3\textwidth}
                \centering
                \includegraphics[width=\linewidth]{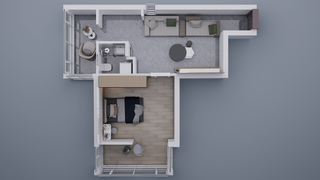}
            \end{subfigure} &
            \begin{subfigure}{0.3\textwidth}
                \centering
                \includegraphics[width=\linewidth]{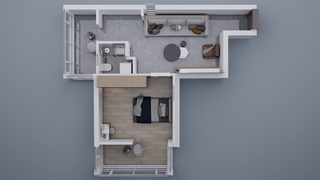}
            \end{subfigure} &
            \begin{subfigure}{0.3\textwidth}
                \centering
                \includegraphics[width=\linewidth]{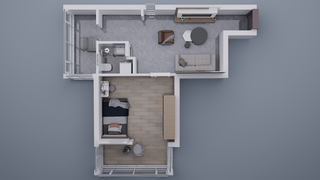}
            \end{subfigure} \\
            \begin{subfigure}{0.3\textwidth}
                \centering
                \includegraphics[width=\linewidth]{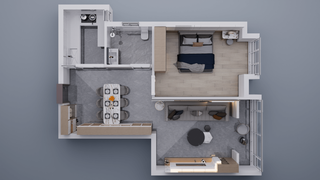}
            \end{subfigure} &
            \begin{subfigure}{0.3\textwidth}
                \centering
                \includegraphics[width=\linewidth]{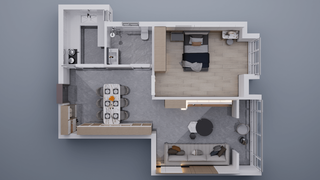}
            \end{subfigure} &
            \begin{subfigure}{0.3\textwidth}
                \centering
                \includegraphics[width=\linewidth]{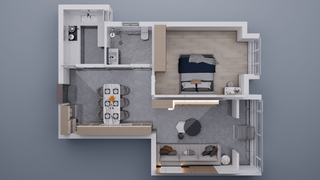}
            \end{subfigure} \\
            \begin{subfigure}{0.3\textwidth}
                \centering
                \includegraphics[width=\linewidth]{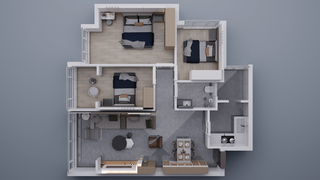}
            \end{subfigure} &
            \begin{subfigure}{0.3\textwidth}
                \centering
                \includegraphics[width=\linewidth]{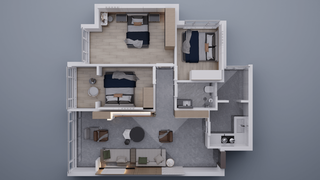}
            \end{subfigure} &
            \begin{subfigure}{0.3\textwidth}
                \centering
                \includegraphics[width=\linewidth]{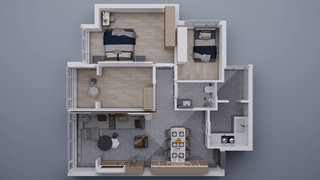}
            \end{subfigure} \\
            \begin{subfigure}{0.3\textwidth}
                \centering
                \includegraphics[width=\linewidth]{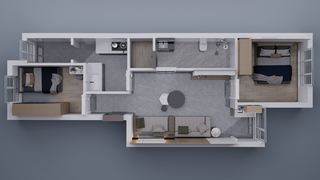}
            \end{subfigure} &
            \begin{subfigure}{0.3\textwidth}
                \centering
                \includegraphics[width=\linewidth]{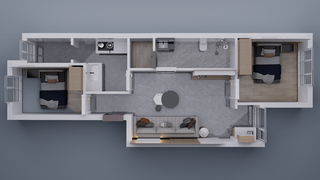}
            \end{subfigure} &
            \begin{subfigure}{0.3\textwidth}
                \centering
                \includegraphics[width=\linewidth]{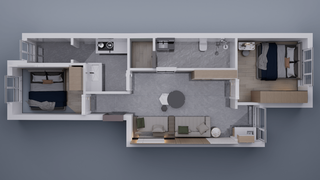}
            \end{subfigure} \\
            \begin{subfigure}{0.3\textwidth}
                \centering
                \includegraphics[width=\linewidth]{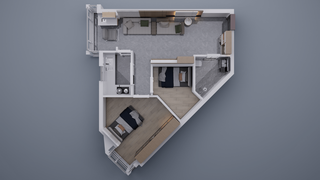}
            \end{subfigure} &
            \begin{subfigure}{0.3\textwidth}
                \centering
                \includegraphics[width=\linewidth]{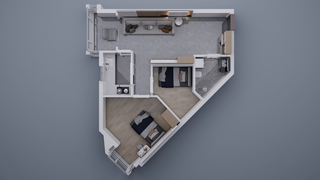}
            \end{subfigure} &
            \begin{subfigure}{0.3\textwidth}
                \centering
                \includegraphics[width=\linewidth]{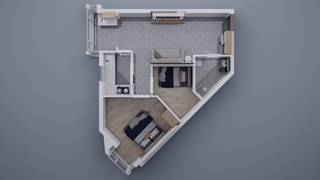}
            \end{subfigure} \\
        \end{tabular}
    }
    \caption{Additional photorealistic rendering of diverse synthesized layouts by CHOrD conditioned on floor plans.}
    \label{fig:layout}
\end{figure}

We present additional qualitative results for fine-grained layout synthesis in Figure~\ref{fig:table_grids}, text-conditioned floor planning in Figure~\ref{fig:text_cond_gen_exp}, open-plan-conditioned floor planning in Figure~\ref{fig:frame_gen}, and photorealistic rendering of floor plan-conditioned layout synthesis in Figure~\ref{fig:layout}.

\begin{figure}
    \centering
    \begin{tabular}{>{\centering\arraybackslash}m{0.3\linewidth} >{\centering\arraybackslash}m{0.3\linewidth} >{\centering\arraybackslash}m{0.3\linewidth}}
        \begin{subfigure}{\linewidth}
            \centering
            \includegraphics[width=\linewidth]{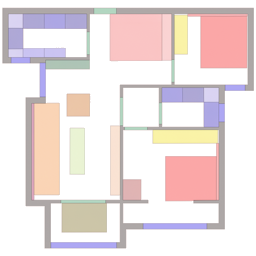}
            \caption*{Unet output}
        \end{subfigure} &
        \begin{subfigure}{\linewidth}
            \centering
            \includegraphics[width=\linewidth]{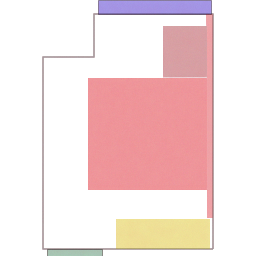}
            \caption*{Unet output}
        \end{subfigure} &
        \begin{subfigure}{\linewidth}
            \centering
            \includegraphics[width=\linewidth]{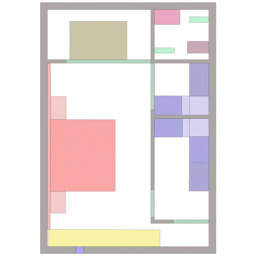}
            \caption*{Unet output}
        \end{subfigure} \\
        \begin{subfigure}{\linewidth}
            \centering
            \includegraphics[width=\linewidth]{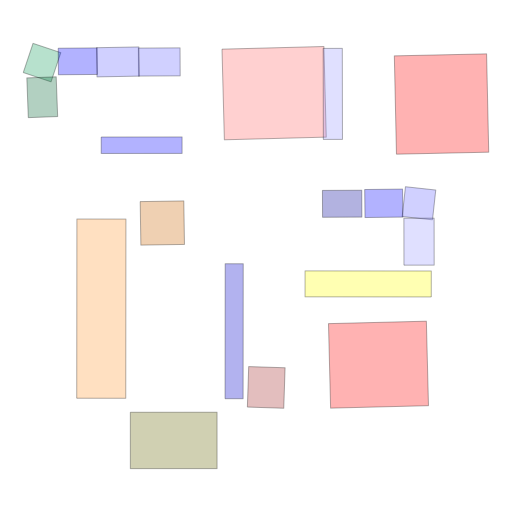}
            \caption*{Kitchen cabinet detection errors}
        \end{subfigure} &
        \begin{subfigure}{\linewidth}
            \centering
            \includegraphics[width=\linewidth]{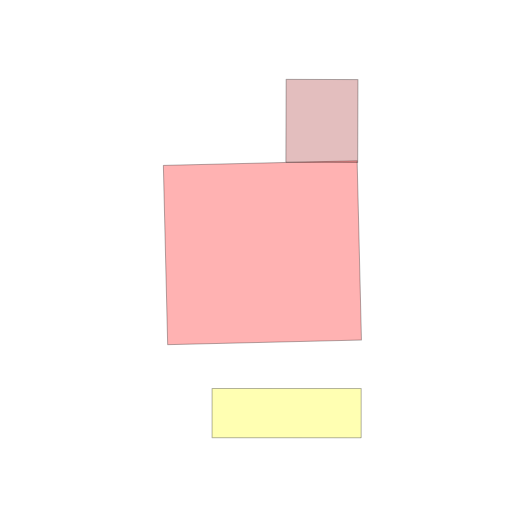}
            \caption*{Bed and table detection errors}
        \end{subfigure} &
        \begin{subfigure}{\linewidth}
            \centering
            \includegraphics[width=\linewidth]{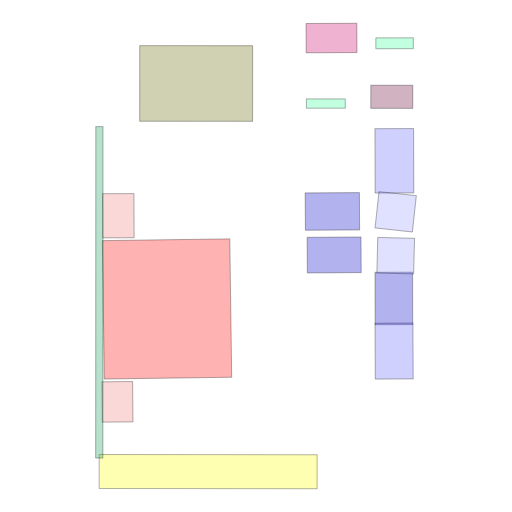}
            \caption*{Kitchen cabinet detection errors}
        \end{subfigure} \\
    \end{tabular}
    \caption{Sporadic failure cases due to YOLO detection errors when trained with insufficient data.}
    \label{fig:failure_cases}
\end{figure}

\begin{figure}
    \centering
    {
    \begin{tabular}{cccc}
        \begin{subfigure}{0.20\linewidth}
            \centering
            \includegraphics[width=1.2\linewidth]{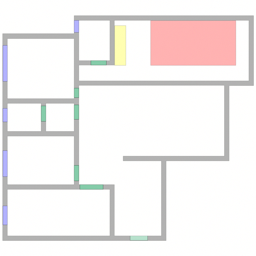}
        \end{subfigure} &
        \begin{subfigure}{0.20\linewidth}
            \centering
            \includegraphics[width=1.2\linewidth]{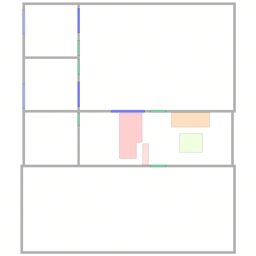}
        \end{subfigure} &
        \begin{subfigure}{0.20\linewidth}
            \centering
            \includegraphics[width=1.2\linewidth]{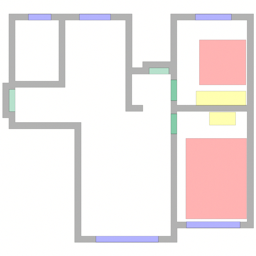}
        \end{subfigure} &
        \begin{subfigure}{0.20\linewidth}
            \centering
            \includegraphics[width=1.2\linewidth]{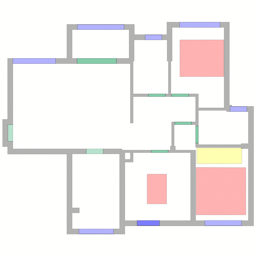}
        \end{subfigure}  \\
        \multicolumn{4}{c}{3D-FRONT} \\
        \begin{subfigure}{0.20\linewidth}
            \centering
            \includegraphics[width=1.2\linewidth]{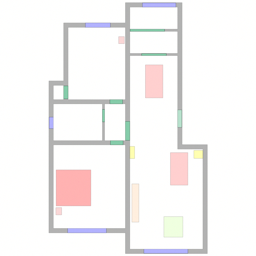}
        \end{subfigure} &
        \begin{subfigure}{0.20\linewidth}
            \centering
            \includegraphics[width=1.2\linewidth]{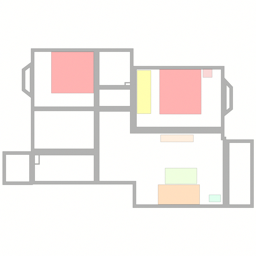}
        \end{subfigure} &
        \begin{subfigure}{0.20\linewidth}
            \centering
            \includegraphics[width=1.2\linewidth]{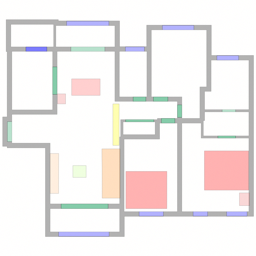}
        \end{subfigure} &
        \begin{subfigure}{0.20\linewidth}
            \centering
            \includegraphics[width=1.2\linewidth]{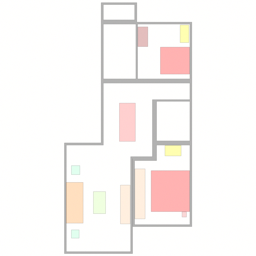}
        \end{subfigure}\\
        \multicolumn{4}{c}{\ourdataset{}} \\
    \end{tabular}}
    \caption{Performance of \ourmethod{} on 3D-FRONT and \ourdataset{}, where results obtained from training on 3D-FRONT exhibit implausible unfurnished rooms due to artifacts in the original database. 
    }
    \label{fig:scene_synthesis}
\end{figure}

In rare instances, YOLOv8 struggled to detect accurate bounding boxes, resulting in misaligned objects or minor collisions, even though the layout images were axis-aligned and collision-free, as shown in Figure~\ref{fig:failure_cases}.
We demonstrated that this can be straightforwardly addressed with more training data.
Specifically, we trained \ourmethod{} on a privately collected dataset of over 100,000 indoor scenes, achieving significantly better results (\textbf{FID} 17.76, \textbf{KID} 0.02, \textbf{POR} 0.005, \textbf{PIoU} $4.399 \times 10^{-5}$) with substantially fewer failure cases compared to the results obtained from training on the \ourdataset{} (9,706 scenes) and reported in Table~\ref{tab:layout_comparison}.
\ourmethod{} also performs considerably better when trained on \ourdataset{} compared to 3D-FRONT, as illustrated in Figure~\ref{fig:scene_synthesis}. These results evidence the strong scalability of \ourmethod{}.

\label{sec:additional-res}

\end{CJK*}



\end{document}